\pdfoutput=1

\documentclass[11pt]{article}

\usepackage[preprint]{acl}

\usepackage{times}
\usepackage{latexsym}

\usepackage{graphicx}

\usepackage{url}            
\usepackage{booktabs}       
\usepackage{amsfonts}
\usepackage{nicefrac}       
\usepackage{microtype}      
\usepackage{lipsum}         
\usepackage{graphicx}
\usepackage{doi}
\usepackage{amsmath}
\usepackage{amssymb}
\usepackage{amsthm}
\usepackage{bbding}
\usepackage{multirow}
\usepackage{subcaption}
\usepackage{algorithm}
\usepackage{algorithmic}
\usepackage{cleveref}       
\usepackage{enumitem}
\usepackage{makecell}       
\usepackage{color}
\usepackage{hyperref}       
\usepackage{minitoc}
\usepackage{adjustbox}

\usepackage[many]{tcolorbox}  
\newtcolorbox{boxRound}{
    boxrule = 1.5pt,
    rounded corners,
    arc = 5pt
}

\usepackage{array} %
\usepackage{colortbl}
\usepackage{pifont}%
\newcommand{\cmark}{\ding{51}}%

\newcolumntype{a}{>{\columncolor{gray!10}}c}

\renewcommand{\paragraph}[1]{\vspace{0.1em}\noindent\textbf{#1}}

\title{Panza: Design and Analysis of a Fully-Local \\ Personalized Text Writing Assistant }

\author{
  Armand Nicolicioiu\thanks{Equal contribution.}\thanks{Institute of Science and Technology Austria.} 
   \And
  Eugenia Iofinova\footnotemark[1]\footnotemark[2]
     \And
   Andrej Jovanovic\footnotemark[1]\footnotemark[2]
     \And
  Eldar Kurtic\footnotemark[1]\footnotemark[2]
  \\ \AND
  Mahdi Nikdan\footnotemark[1]\footnotemark[2]
  \And
  Andrei Panferov\footnotemark[2]\\
  \And
  Ilia Markov\footnotemark[2]\\
  \And
  Nir Shavit\thanks{Massachusetts Institute of Technology.}\\
  \And
  Dan Alistarh\thanks{Correspondence to \texttt{dan.alistarh@ista.ac.at}}\footnotemark[2]\\
}

\begin{document}
\maketitle

\begin{abstract}
    The availability of powerful open-source large language models (LLMs) opens exciting use-cases, such as using personal data to fine-tune these models to imitate a user's unique writing style. Two key requirements for such assistants are \emph{personalization}–in the sense that the assistant should recognizably reflect the user's own writing style—and \emph{privacy}–users may justifiably be wary of uploading extremely personal data, such as their email archive, to a third-party service.
In this paper, we present a new design and evaluation for such an automated assistant, for the specific use case of email generation, which we call \emph{Panza}.
Panza's personalization features are based on a combination of fine-tuning using a variant of the Reverse Instructions technique~\citep{koksal2023longform} together with Retrieval-Augmented Generation (RAG). We demonstrate that this combination allows us to fine-tune an LLM to reflect a user's writing style using limited data, while executing on extremely limited resources, e.g. on a free Google Colab instance. 
Our key methodological contribution is the first detailed study of \emph{evaluation metrics} for this personalized writing task, and of how different choices of system components--the use of RAG and of different fine-tuning approaches--impact the system's performance. Additionally, we demonstrate that very little data - under 100 email samples - are sufficient to create models that convincingly imitate humans. This finding showcases a previously-unknown attack vector in language models - that access to a small number of writing samples can allow a bad actor to cheaply create generative models that imitate a target's writing style. 
We are releasing the full Panza code as well as three new email datasets licensed for research use at \url{https://github.com/IST-DASLab/PanzaMail}.
\end{abstract}

\section{Introduction}
An automated personal assistant is a software application that can help a user with various repetitive tasks such as email, writing, or summarization. 
Large Language Models (LLMs) are natural candidates for implementing personal assistants, as they can provide remarkably good results on such generative tasks. At the same time, many highly capable LLMs can only be accessed via an API, either because their weights are proprietary, or because the model size makes self-hosting and customized finetuning cost-prohibitive. This makes it expensive or impossible to support certain natural features for automated personal assistants, namely:

\begin{enumerate}
\setlength{\itemsep}{2pt}
  \setlength{\parskip}{2pt}
  \setlength{\parsep}{2pt}
    \item \textbf{Model personalization}: customizing the model to the individual's tone, preferences, and personal history;
    \item \textbf{Privacy protection}: allowing the model to have access to highly personal information of a caliber that---unlike corporate data---most people would not agree to share, even if promised that the data is protected by the provider’s cloud.
\end{enumerate}

A natural approach to addressing these constraints would be to \emph{execute these models locally, on the user's own data and hardware}. 
However, this poses obvious challenges both in terms of \emph{data ingestion} -- getting user data into a format that can be used to successfully train or fine-tune an LLM -- and in terms of \emph{hardware efficiency} -- specifically, because fine-tuning or even running inference over a capable LLM on user data may be technically impossible if done naively. 

In this application-driven paper, we focus on the challenging case of designing and implementing a \emph{fully-local personalized LLM-powered email writing assistant}, which we call Panza. Panza's goal is to generate messages in the user's style, given a short user prompt and, optionally, access to a set of previous emails. We demonstrate that it is possible to obtain a capable assistant starting from relatively-small existing pre-trained models such as Llama-3-8B~\citep{llama3} or Phi-3-Mini \citep{abdin2024phi3technicalreporthighly}, and that these models can be trained on a machine with a commodity GPU, or even CPU-only, and run on a commodity laptop, such as a MacBook. The overall structure of Panza is illustrated in Figure~\ref{fig:panza-diagram}. 
While the focus of our work is application-driven, we present interesting findings from the following research perspectives:

\begin{figure}[t]
    \centering
    \includegraphics[width=0.46\textwidth, height=3.5cm]{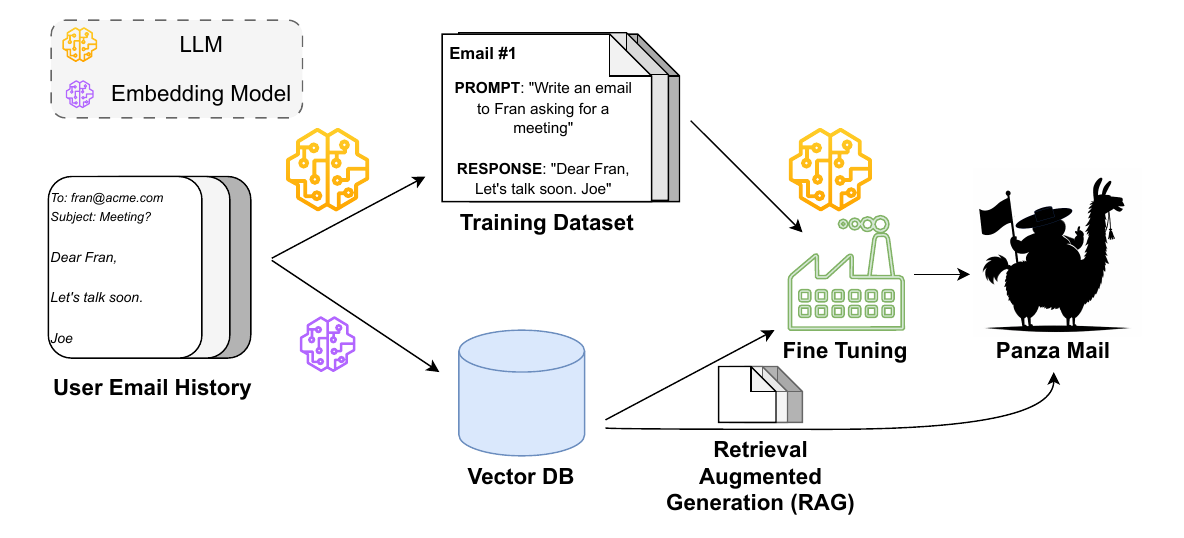}
    \vspace{-6pt}
    \caption{Panza's overall design. 
    Given a set of emails produced by the user, we produce both a finetuning dataset (using Reverse Instructions) and retrieval augmented generation (RAG) database. 
    The base model is first fine-tuned and then served  with RAG.}
    \label{fig:panza-diagram}
\end{figure}

\begin{enumerate}
	\item We extend the Reverse Instructions algorithm of~\citet{koksal2023longform} to the case of personalizing a generative LLM's output to match a user's writing style. The original goal of Reverse Instructions was to convert unlabeled text data to instruction data by using an LLM to output a synthetic instruction that would result in the original data. We extend this idea by applying it to the personalization problem, by demonstrating its effectiveness on much smaller data sizes, and by combining it with Retrieval-Augmented Generation (RAG) \citep{lewis2020retrieval}, used to retrieve relevant emails at inference time to provide additional context.  

	\item We perform an extensive analysis of the \emph{text personalization task} in this context. We begin with an investigation of evaluation metrics, in conjunction with studying the impact of different design options, training, and deployment techniques on the final model performance. First, we show that the traditional BLEU / ROUGE / BERT scores for measuring text generation quality show similar trends, but \emph{fail to fully capture performance on this task}. However, we observe that the \emph{combination of BLEU and MAUVE scores} provides very strong correlations with human preferences across various user studies.  
 
     \item With these metrics in place, we show that \emph{fine-tuning on the user's data} consistently outperforms a pre-trained instruction-tuned model, even when augmented with RAG. Further, we perform one of the first in-depth studies of the impact of RAG on \emph{personalized} model performance when applied at training time or at inference time for long-form generation.

	\item We show that this entire pipeline can be executed in a \emph{resource-constrained setup}. 
	Both the fine-tuning and the inference / RAG components can be run efficiently and accurately on a system with a single commodity GPU, e.g. a free-tier Google Colab instance. The best parameter-efficient fine-tuning (PEFT) results are obtained using the Robust Adaptation (RoSA) method~\citep{nikdan2024rosa}, which we find particularly suited for style transfer. Our main innovation here is new and accurate adapter merging,  required to combine sparse and low-rank PEFT adapters into quantized weights. Additionally, we demonstrate the practical feasibility of using Panza on a personal computer through a combination of~\citet{ollama} and a new Chrome plugin which integrates Panza into GMail. 

    \item Finally, we highlight the \emph{emerging risks} of open-source LLMs by demonstrating that as few as 50-100 emails suffice to make a convincing imitation of a person's writing style. This uncovers a novel attack vector, where malicious actors can use leaked data to create credible---and difficult-to-trace---imitations of the breach victims' writing style.
\end{enumerate}

These findings are confirmed via a range of human user studies, showing that, using our techniques, finetuning existing open LLMs achieves high-quality personalized text generation, and that the resulting models are both sufficiently convincing and sufficiently differentiated, in the sense that humans can attribute the resulting Panza outputs to the correct (human) author.

\paragraph{Related Work.}
The problem we consider is that of customizing an LLM to yield outputs that match a given user's tone and writing style. 
Additional challenges are the limited availability of user data, and the stringent computational constraints. 
The problem of personalization has seen a lot of interest for Language Models, e.g.~\citet{king2020evaluating}. 
In addition, the recent explosion of highly performant models of the 4-8B parameter scale has made individualized personalization a viable goal~\citep{chen2023large, kirk2024benefits, zhang2024personalizationlargelanguagemodels}. To organize these efforts, the Language Personalization benchmark~\citep{lamp} was recently introduced; however, it consists only of classification and summarization-style tasks, and does not introduce any benchmarks for long-form text generation.

Our work builds upon the Reverse Instructions algorithm~\citep{koksal2023longform}, which allows us to convert long-form user data into training data, by using an LLM to write synthetic prompts for each email. However, this work did not consider personalization; we precisely contrast the two approaches in Appendix~\ref{sec:longform_comparison}, showing that the personalized approach of Panza achieves much better performance than the approach of~\citet{koksal2023longform}.

We show both via examples and metrics comparisons that both finetuning and RAG are helpful for the task: while base LLMs with RAG can provide coherent outputs, including personal information obtained via retrieval, they largely fail to match a given user's writing style. A similar trend has been observed in OPPU~\citep{tan2024OPPU}, which also deals with personalized finetuning, but they investigate this problem in the context of text condensation,  such as article titling and text paraphrasing, while Panza tackles the harder problem of long-form text generation from a shorter prompt. In addition, we consider the practical problem of computing requirements, and investigating the effects of parameter-efficient finetuning methods (LoRA~\citep{hu2021lora}, RoSA~\citep{nikdan2024rosa}) and of quantization.

We note briefly other works proposing variants of Reverse Instructions.  \citet{sennrich-etal-2016-improving} uses generated prompts for machine translation, but this paper does not deal with long-form text generation.  Instruction Backtranslation~\citep{li2024backtranslation} contains a step in which an LLM is used to construct a prompt that would result in the generation of an unlabeled document; this paper likewise does not deal with personalization. Further, unlike the findings in that paper, we demonstrate through a user study that the combination of Reverse Instructions and RAG is sufficient to provide high-quality emails in a user's style, without requiring the additional filtering steps of Instruction Backtranslation---which would anyway be impractical for the email generation task, due to the very limited data size.

\section{Method}

The Panza design, described in Figure~\ref{fig:panza-diagram}, requires a pre-trained, instruction-tuned LLM and a set of emails sent by the user.
Both the LLM and the emails have dual use.
First, the pre-trained LLM is used to rephrase the ``raw'' user emails in the form of instructions, which will be used for finetuning. 
Then, the LLM itself is going to be fine-tuned on these instructions, resulting on the Panza personalized model. 
Independently, emails are integrated in a RAG database, used at runtime. 

\subsection{Finetuning via Reverse Instructions}

Finetuning is the core personalization technique behind Panza. The task data is typically a set of several hundred or thousand emails in the user's history. We convert these emails to training data by a variant of Reverse Instructions.
The intuition is that we would like to ``reduce'' the problem of generating emails in the user's style to a specific instance of instruction-tuning: in practice, the user query would come in the form of an instruction, e.g. ``Write an email to Joanne to set up a meeting on Wednesday.'', and the ``correct'' answer would be such an email written in the user's style. To induce this behavior from the LLM, our training works in two steps:
    First, we use a pre-trained LLM to rephrase each email into a targeted instruction, containing just enough information for an assistant to write the original email. See Appendix~\ref{sec:appendixPrompting} for the prompt we use for this task.
    
    Then, we fine-tune the backbone LLM on the (instruction, email) pairs obtained in the first step, with the standard next-token prediction training objective, which induces the LLM to reconstitute the original email as a response to the instruction. 
    Optionally, we implement a training-time RAG component, which retrieves query-related emails (via cosine similarity of the emails and query in the embedding space of a smaller LLM), and provides them as context to the LLM at training time. This  is similar to retrieval-augmented fine-tuning (RAFT)~\citep{zhang2024raft}. 
At the end of these two steps, we have obtained a personalized LLM which can respond to short user queries by writing targeted emails that should follow the user's style.

\subsection{Local Fine-Tuning}

In the absence of computational or memory constraints, we perform full fine-tuning (FFT) of the base model, 
and inference over it, possibly adding a retrieval-augmented generation (RAG) component, which retrieves similar emails sent in the past. 
However, fine-tuning and deploying a powerful billion-parameter model locally requires a powerful GPU with significant memory. Therefore, we investigate techniques for reducing these costs, as well as their impact in terms of accuracy metrics. 

Memory efficiency is critical in our setting. For instance, full fine-tuning (FFT) of a \texttt{Mistral-7B} model \citep{jiang2023mistral} in half-precision using a standard Adam optimizer \citep{kingma2014adam} requires more than \texttt{60GB} of GPU memory, which is not available on a consumer-grade  machine.
We tailor Panza to two resource-constrained settings: running on a GPU with under \texttt{24GB} RAM,  and under \texttt{15GB} RAM. Throughout, we use \texttt{Mistral-7B} \citep{jiang2023mistral} as a running example for costs, but the techniques apply to other base LLMs with similar size. We always consider a local training micro-batch size of $1$ to minimize memory footprint, accumulating gradients whenever necessary. See Section \ref{sec:results} for more details. Training usually takes under an hour.

\paragraph{Finetuning on a single GPU.}
We first assume access to a single \texttt{24GB} GPU (such as NVIDIA GeForce RTX 3090), i.e. a GPU desktop or a strong gaming laptop. 
In this case, we use Parameter-Efficient Fine-Tuning (PEFT) methods, which tune only a small set of parameters to enable adaptation of models to downstream tasks. We  investigate the standard LoRA method~\citep{hu2021lora}, and the more recent Robust Adaptation (RoSA) method \citep{nikdan2024rosa}, which we find to be particularly effective for style transfer. Specifically, by training a combination of low-rank and sparse adapters on top of the base weights, RoSA allows effective fine-tuning of a half-precision \texttt{8B} model on less than \texttt{24GB} of memory, with competitive accuracy relative to full fine-tuning. For deployment, we merge the RoSA adapters into the base model weights, with the inference requiring around \texttt{15GB}. In Appendix~\ref{appendix:panza-quant} we demonstrate the feasibility of training Panza in under 15GB of memory by additionally incorporating quantization, including a detailed study of quantization approaches.

\subsection{Inference}
If the GPU is local to the user, for instance on a gaming laptop, then the same hardware can simply be used to run inference on the model at any time. If, however, the model is trained on external hardware (for instance, on Google Colab, Lightning Studios, or another cloud provider), using the same hardware for inference becomes impractical: keeping the hardware running with the model preloaded is expensive, while loading the model onto the GPU every time Panza is called is too slow to be practical. Instead, we convert the model to a \texttt{.cpp} version that runs on CPU via Ollama~\citep{ollama}. Using this setup, we are able to run the model on an M-Series MacBook Pro (M3 Pro Chip with 18 GB of memory), achieving $16.09$ tokens per second, or about $6.21$ seconds to generate an average email. To further demonstrate the potential for a positive user experience, we built a sample Google Chrome extension that incorporates Panza into GMail (see Appendix~\ref{appendix:google_chrome_extension}).

\section{Evaluation Protocol}

\subsection{Datasets}

A key challenge of this project is the lack of availability of personal e-mail datasets, due to the sensitive content of the data. To our knowledge, the only available repository contains the business e-mails of 144 Enron employees, originally released by the Federal Energy Regulatory Committee; the version we use~\citep{enron} is licensed for research. We use the emails of four employees (Sara Shackleton, Kay Mann, Jeff Dasovich, and Tana Jones) with over 400 (English) e-mails each, hereafter identified by their first names. To avoid influencing the model by explicitly invoking Enron, \emph{we changed the names of the corporation and its executives}. 

We use five additional datasets. The ``David'', ``Isabel'', and ``Marcus'' datasets were anonymized manually by the authors and donated for research use by volunteers with a clear understanding of its release and proposed use; these datasets will be released as part of this project. Two additional datasets, identified as Anon1 and Anon2, were also donated by volunteers for this project, but will not be released due to the sensitive nature of their contents. Of the nine datasets used, six contain primarily business emails, and three contain a mix of business and personal emails.

\begin{table}[t]
\centering
\caption{E-mail datasets used. The David, Isabel, and Marcus datasets are published along with this work.}
\label{tab:datasets}
\resizebox{0.75\linewidth}{!}{

\renewcommand{\arraystretch}{1.2}

\begin{tabular}{llcc}%
\hline%
Name&Source&Public& Number of Emails\\%
\hline%
Tana & Enron & \cmark & 742 \\
Jeff & Enron & \cmark & 573 \\
Kay & Enron & \cmark & 481 \\
Sara & Enron & \cmark & 464 \\
David & Donated & \cmark (new) & 166 \\
Isabel & Donated & \cmark (new) & 210 \\
Marcus & Donated & \cmark (new) & 150 \\
Anon1 & Donated &  & 301 \\
Anon2 & Donated &  & 669 \\
\hline%
\end{tabular}%

}
\end{table}

The test-train split for fine-tuning / PEFT is 80-20\%, and only training set emails are used to form the RAG database employed at inference time. 

\subsection{Metrics}

Panza uses the text generation capabilities of LLMs for two tasks: to summarize a user's emails to create synthetic prompts via  Reverse Instructions, and to generate new emails in the test-time prompts. These two use cases both rely on the ability of LLMs to summarize or rephrase content; the email generation task additionally requires the recall of both general and user-specific knowledge, and an imitation of a user's personal style.
Thus, we divide the evaluation benchmarks into four categories - phrasing quality, user-specific knowledge, general knowledge, and style.

\paragraph{Phrasing quality.} For phrasing quality we rely on the  BLEU~\citep{Papineni2002BleuAM} and ROUGE~\citep{Lin2004ROUGEAP} metrics, which are standard to measure translation and summarization quality. Both metrics function by counting matching N-grams between the LLM output and one or several `golden' responses (for email generation, the golden response is the actual user  email). The BLEU score is a weighted measure of 1-4-grams that match exactly between the output and golden text strings, normalized by the string length; we use an equal weight of $0.25$ for each N-gram length. ROUGE reports 1-gram and 2-gram precision, recall, and F1-score, as well as the longest substring precision, recall, and F1-score; in our paper, we use the longest-substring F1 score for maximum contrast with the BLEU metric. For both metrics, we use the Torchmetrics package; they are computed for each prompt/output combination, and we report the average across all prompts as the overall value. We do not compute either metric on a per-sentence basis, but rather compare N-grams in the full email text after dropping punctuation.

\paragraph{User-specific knowledge.} As there is no database of user-specific knowledge, we restrict the evaluation of such knowledge to what is contained in the user's emails. This  working assumption enables us to use RAG-assisted email generation; here, it further enables to evaluate user-specific knowledge on the same test dataset as for the paraphrasing quality. We do not attempt to decouple knowledge content from paraphrasing quality, rather relying on the overall BLEU and ROUGE scores to reflect the correct imputed information. Note that, unlike the general knowledge desideratum, the user-specific information is better specified---a prompt requesting the user's current address is easier to evaluate than one asking for suggestions for a travel destination---and so the N-gram match is an appropriate measure of quality.

\paragraph{General knowledge.} For world knowledge measurements, we rely on the standard six Open-LLM leaderboard evaluation suite~\citep{open-llm-leaderboard}. Together, these tasks test the model's ability to perform basic reasoning and factual recall in English, and are used as a broad evaluation of a model's quality. Overall, Panza models retain over 98\% of the performance of Llama-3-8B-Instruct on this benchmark. Full details are in Appendix~\ref{sec:general_knowledge_extended}.

\paragraph{Style transfer.} To measure the quality of style transfer, we use the MAUVE score \citep{Pillutla2021MAUVEMT}, introduced to measure the degree of similarity between machine-generated and human-written text.  Specifically, the MAUVE score relies on an estimate of the symmetric K-L divergence between the distribution of authentic text and the distribution of model outputs, which we parametrize as standard.  
 
Higher MAUVE score is correlated with higher similarity between the generated text and human text.

\paragraph{Discussion.} A major challenge of this project is that the email generation task differs significantly from summarization or translation tasks, in that we generally expect the output email to be \emph{longer and more detailed} than the prompt used to generate it, thus requiring some improvisation on the part of the model, and making it unlikely for the generated email to be close to the actual one. This is reflected in the BLEU/ROUGE scores, which are  lower than what would the reader may have observed for translation or summarization tasks. Manual review during the project development phase has shown that, nevertheless, these scores appear highly correlated with output email quality.  We provide examples in Appendix~\ref{sec:appendixGenerations}, validate BLEU and MAUVE in a human study in Appendix~\ref{sec:additional_human_studies}, and consider BERT-score and METEOR metrics in Appendix~\ref{appendix:alternate_metrics}.
As a rough guideline, human subjects generally agreed that models achieving above 0.2 average BLEU and 0.75 MAUVE produced plausible emails.

\section{Experimental Results}
\label{sec:results}

\subsection{Baselines}
As no baselines appear to exist for the task of \emph{personalized} email generation, we investigate the effect of finetuning by baselining against simply using prompt engineering to elicit personalization. (We additionally compared to the results of~\citet{koksal2023longform} in Appendix~\ref{sec:longform_comparison} and to GPT-4o prompted with a large subset of the user's entire email history in Appendix~\ref{sec:gpt}.) We start with publicly available instruction-finetuned LLMs: Meta-Llama-3-8B-Instruct, Mistral-7B-Instruct-v0.2, and Phi-3-\-mini-\-4k-\-instruct.
As a first baseline, we prompt models in the following format. 
The system preamble sets the role of the LLM as an email writing assistant, while the user preamble contains general information about the user, such as name, occupation, and address, which has been shown to be essential to personalizing LLMs through prompt engineering \citep{wu2024understandingroleuserprofile}.
    Finally, we provide the actual email writing instruction, as created during the Reverse Instructions step.

This baseline (denoted \textbf{Pretrained}) writes well-structured emails, but without explicit personalization.
Next, we test if presenting a few samples of the user's previous emails during inference through a RAG component can provide enough style information. We select the closest $n_{RAG}$ previous emails, filtered by a relevancy threshold $T_{RAG}$, and add them to the input as an additional preamble. We denote this baseline as \textbf{Pretrained + RAG}.

\subsection{Fine-Tuning via Reverse Instructions}

After generating pairs of (instruction, email), we fine-tune the pre-trained LLM to reconstruct the email given the instruction.
We analyze the following regimes: full fine-tuning (\textbf{FFT}) and two PEFT methods: Robust Adaptation (\textbf{RoSA}) and Low-Rank Adaptation (\textbf{LoRA}). For both training and testing, the input is formatted with the same system and user preambles described for the baselines. We also test whether RAG can bring additional improvements.

Furthermore, we explore if the model can learn how to better leverage previous e-mails in RAG by presenting the same type of augmented prompt during fine-tuning, parameterized by the number of closest emails $n_{RAG}$ and the relevancy threshold $T_{RAG}$. Additionally, to make the model robust to the absence of similar emails, we have a $p_{RAG}$ chance of not using any similar emails for a particular instruction during fine-tuning, even if there are matches in the database. This adapts~\citet{zhang2024raft} to our setting and is denoted as \textbf{RAFT} (Retrieval-Augmented Fine-Tuning).

\paragraph{Hyperparameter tuning.} We found  fine-tuning, and especially PEFT, to be highly sensitive to learning rate and the number of training epochs. To find suitable hyperparameters, we first used a greedy grid search approach with learning rates ranging from $10^{-6}$ to $10^{-4}$ and epoch ranges from $1$ to $9$, batch sizes of $8$ and $16$, and  using the BLEU metric as the proxy for overall model quality. We used the realistic, non-anonymized Anon1 and Anon2 datasets for hyperparameter tuning and chose values that worked well for both. Overall, we found that learning rate of $10^{-5}$, batch size of $8$, and 5 epochs (3 for FFT) to work well across all base models and finetuning styles. We then did limited tuning of these parameters for the other users. The values for all users are presented in Appendix~\ref{sec:appendixHyperparameters}.

\begin{figure}[t!]
\centering
\begin{minipage}{.45\textwidth}
  \centering
  
  \includegraphics[width=.95\linewidth]{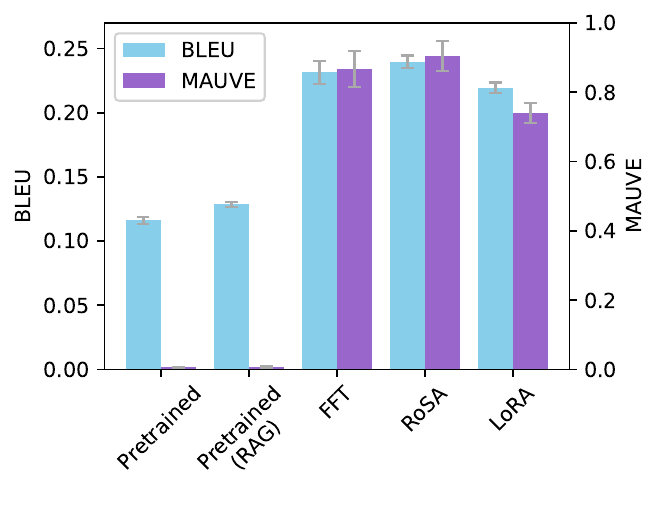}
  \vspace{-20pt}
  \captionof{figure}{Average performance of finetuning-based methods compared to pretrained baselines (Llama3-8B-Instruct with and without RAG at inference). All versions of finetuning (FFT, RoSA and LoRA) outperform the baselines; RoSA and FFT outperform LoRA.}
  \label{figRegimes}
\end{minipage}%
\end{figure}

\subsection{Results Across Methods and Models}

We find that all fine-tuning regimes outperform the Pretrained + RAG baselines by a large margin, adapting to the user's writing style. The results are illustrated in Figure~\ref{figRegimes}; we present a qualitative comparison of the generated emails in Appendix~\ref{sec:appendixGenerations}; using leading closed LLMs yielded similarly poor results. RoSA performs on par with FFT, consistently surpassing LoRA, especially in terms of MAUVE score. The same trend is observed for all the backbones we trained, and for all users (see Appendix \ref{sec:appendixFinetuning} for full results).

\begin{figure}[t!]
\begin{minipage}{.45\textwidth}
  \centering
  
  \includegraphics[width=.95\linewidth]{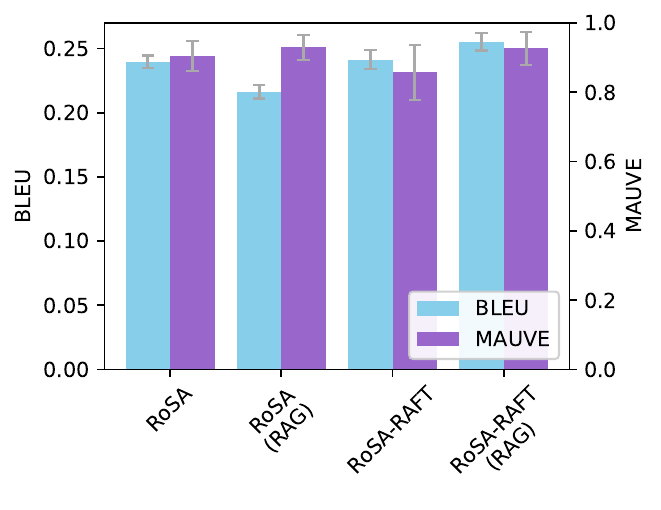}
  \vspace{-20pt}
  \captionof{figure}{Comparison between instruction-only fine-tuning and Retrieval-Augmented Fine-Tuning (RAFT) for RoSA on Llama3-8B, averaged across users.
  }
  \label{figRAG}
\end{minipage}
\end{figure}

In Figure~\ref{figRAG} we show the effect of RAG on models fine-tuned with RoSA. We observe that, although RAG clearly improves over the pretrained baseline, it reduces the average BLEU score for the RoSA fine-tuned model (second group), but may slightly increase the average MAUVE score. 
This is not desirable, as BLEU score (relative to the ground-truth email) is a closer measure of content accuracy than MAUVE. 
Based on analyzing individual samples, we hypothesize that this may be caused by the model re-using the RAG context too aggressively at deployment time, leading to emails that are very similar to past emails--preserving style--but less accurate in terms of content, leading to lower BLEU. We  overcome this issue by introducing RAG during fine-tuning, i.e. using RAFT:  the model can then ``learn'' to ignore irrelevant emails present in the RAG context.

\subsection{Style Evaluation}

\begin{figure}[t!]
\centering
\begin{minipage}{.45\textwidth}
  \centering
  
  \includegraphics[width=.95\linewidth]{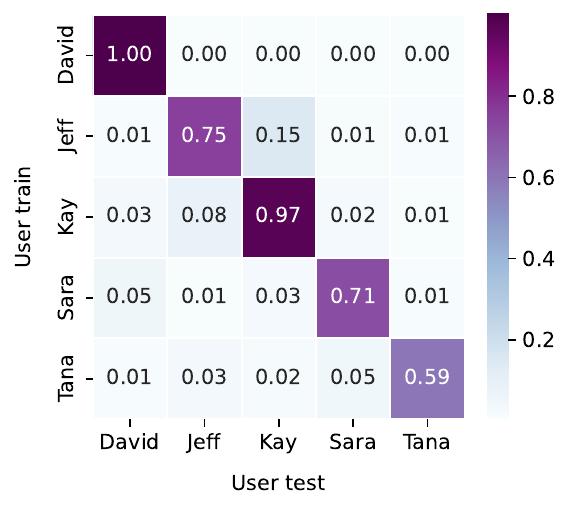}
  
  \captionof{figure}{Style comparison between models trained for different users. Each model, trained for a particular user, is used to generate emails for unseen instructions of all the users. We measure the MAUVE score between the generations and the original emails written by the user.}
  \label{figConfusionMatrix}
\end{minipage}%
\end{figure}

Recall that MAUVE measures the gap between machine-generated text and human-written text. Above, we reported the MAUVE scores on the test emails coming from the same user the model was trained for. Next, we do a pairwise comparison, evaluating models trained for different users on the test data of all the other users. This focuses precisely on how well style is reflected in generated emails. 
In Figure~\ref{figConfusionMatrix}, we see that each model produces a high MAUVE score $(0.6$ to $1.0)$ \emph{only for the test emails of the user it was trained for}, while it has close to $0$ MAUVE score on any other user. In terms of BLEU/ROUGE score, all models have essentially the same performance on the test set of any given user, suggesting all models have similar paraphrasing capacity to express the given instruction, but each does it in the style of the user it was trained for. 

\subsection{Impact of Training Dataset Size on Model Quality}\label{sec:ablation}
\begin{figure}
\centering
\begin{minipage}[c]{\linewidth}
\centering
   \includegraphics[width=0.9\linewidth, height=19em]{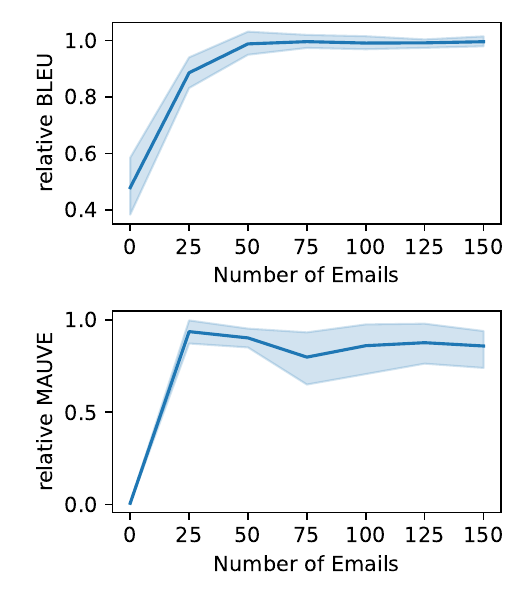}
\end{minipage}
\caption{BLEU and MAUVE scores of Panza models trained on smaller subsets of the training data, relative to the maximum score attained for the dataset. 0 emails corresponds to the un-finetuned Llama3-8B-Instruct model. Scores are averaged across seven users.}
    \label{fig:nemail_ablation}
\end{figure}

The success of Panza on the test datasets naturally leads to the question: how little data is needed to create a convincing Panza model? While restricting data size enables further efficiency gains, this functionality also opens up avenues of attack for malicious actors, as they would need access to a small number of emails to successfully imitate a victim. 

We performed an ablation study on the number of emails needed to train a Panza model. We trained individual Panza models on the users \texttt{david}, \texttt{isabel}, \texttt{marcus}, \texttt{jeff}, \texttt{kay}, \texttt{tana}, and \texttt{sara}, varying the number of emails in the training set from $25$ to $150$, using RoSA, and compare that to Panza performance when trained on the full available training data (without RAFT or RAG). In order to compensate for the small data size, we increase the learning rate to $0.0001$ for dataset sizes of $25$ and $50$, and we train models of e-mail size $25$ and $75$ for $11$ epochs and models of $50$ and $100$ for $9$ epochs (however, the $75$- and $100$-email models use the standard learning rate of 1e-5). 

We present the aggregate BLEU and MAUVE metrics for this study in Figure~\ref{fig:nemail_ablation}. In order to meaningfully aggregate between users, we normalize each user's BLEU and MAUVE scores by those obtained by the model trained on the maximum available number of train emails (80\% of the total dataset size). These results show that at a training set of 25 we already observe average BLEU scores at 90\% of the maximum and MAUVE scores of about 0.85, suggesting that the resulting PANZA models are picking up many features of the user's style; by roughly 75-100 emails, the BLEU scores essentially match those of the full training sets, although MAUVE scores take longer to saturate. We therefore conclude that even quite small datasets may be sufficient to successfully impersonate a user. We provide a human study testing this conclusion in Section~\ref{sec:impersonation_human_eval}.

\section{User Evaluation}
In order to further evaluate the efficacy of Panza in generating effective email personalisation, we create three types of user studies: i) \textbf{usefulness} -- to assess whether Panza models provide practical benefit for the user in their email generation, ii) \textbf{persona recognition} -- to assess whether well-performing Panza models are successful in representing the authors' style, and iii) \textbf{impersonation} -- to assess whether Panza produces emails that sound believably human. The details of each study's setup are presented in full in Appendix \ref{appendix:full_user_study_details}.
\subsection{Usefulness}
\label{sec:usefulness}

\begin{table}[b]
\centering
\caption{Human evaluation of Panza usefulness.}
\label{tab:usefulness_eval}
\small{
\setlength{\tabcolsep}{1pt}
\resizebox{\linewidth}{!}{
\begin{tabular}{lcc|aa|cc|aa|cc|aa|c}
\toprule
 & \multicolumn{2}{c}{Subject 1} & \multicolumn{2}{a}{Subject 2} & 
 \multicolumn{2}{c}{Subject 3} &
 \multicolumn{2}{a}{Subject 4} &
 \multicolumn{2}{c}{Subject 5} & \multicolumn{2}{a}{Subject 6} & \\
\cmidrule(lr){2-3}\cmidrule(lr){4-5} \cmidrule(lr){6-7} \cmidrule(lr){8-9} \cmidrule(lr){10-11} \cmidrule(lr){12-13}
 & \makecell{Fixed} & \makecell{Own}  & \makecell{Fixed}  & \makecell{Own} & \makecell{Fixed} & \makecell{Own} & \makecell{Fixed} & \makecell{Own} & \makecell{Fixed} & \makecell{Own} & \makecell{Fixed} & \makecell{Own} & Avg. \\
\midrule

Llama + RAG & 2.00 & 2.13 & 2.25 & 2.25 & 2.38 & 2.69 & 1.38 & 1.19 & 1.69 & 1.50 & 2.94 & 2.81 & 2.10 \\
RoSA + RAG & 2.50 & 2.56 & 2.13 & 2.81 & 2.56& 2.81 & 2.94 & 2.81 & 2.88 & 2.88 & 2.50 & 2.38 & 2.65 \\
FFT + RAG & 2.63 & 2.75 & 2.00 & 2.75 & 1.81 & 2.62 & 2.88 & 2.94 & 2.69 & 2.69 & 2.19 & 2.75 & 2.56 \\
\bottomrule
\end{tabular}
}
}
\end{table}

To assess practical usefulness, we conducted a study in which six subjects were recruited and asked to train Panza models on their own data. 
The subjects were instructed to to train two Panza models, following the standard FFT-RAG and RoSA-RAG setups with no hyperparameter tuning. They then used these models to generate responses to 16 fixed prompts, and 16 prompts drawn from the subject's test set. The subjects then assigned each generated email a rating of 3 (Very useful), 2 (Moderately useful), or 1 (Not useful). Full details are available in Appendix~\ref{appendix:usefulness_study}. 

The results are presented in Table~\ref{tab:usefulness_eval}. We observe that, while on average the outputs of all models were rated as somewhat useful or better, the outputs of Panza models were generally found to be more useful than those of Llama + RAG, with average scores of 2.10 for Llama, 2.56 for FFT, and 2.65 for RoSA. We further note that for FFT and Llama, we see substantially better performance on the rater's own prompts than the fixed set, which may be seen as evidence that Panza tailors to the email \emph{content} distribution of the user's prompts.
\subsection{Persona Recognition}

The intention of this study is to assess whether well-performing Panza models are successful in credibly representing the authors' style. We conducted three separate trials with four participants. Each trial is conducted with a group of four participants. In each trial, two individuals served as the authors, and were asked to train Panza models on their personal emails, and two additional evaluators who served as evaluators. Crucially, all participants in each trial were personally acquainted with both authors. Each participant was then asked to match the e-mails produced by four Panza models to their source data. Two of the four models were trained by the authors on their data. The third model was finetuned on the \texttt{Jeff} dataset, and the final model is an un-finetuned Llama-3-8B-Instruct model augmented with RAG. We note that none of the subjects in this study is personally familiar with \texttt{Jeff}; we included this additional control to to further validate that the subjects are not just distinguishing finetuned from pretrained model output.

For each evaluation task, evaluators were given 24 prompts and the four email responses from the four different writers in a random order, and asked to attribute these four output emails to the four model versions. The prompts to generate the responses were taken from a curated list of custom prompts.

The results are presented in Table~\ref{tab:persona_recognition}, averaged over the three independent trials. We observe that the subjects were able to identify Llama and Panza-Jeff outputs with high accuracy, and were able to identify Subject 1 and Subject 2 from the other Panzas with roughly 64\% precision and recall, far higher than random guessing (25\%).

The greatest source of error was attributing the emails of Subject 1 to Subject 2 or vice versa, which may be due to the fact that each pair of authors are mutual acquaintances, which may indicate increased stylistic similarity. Overall, this study supports our hypothesis that finetuning with Panza creates distinguishable personas.

\begin{table}[t]
    \caption{Persona recognition evaluation: Precision (P) and recall (R) of Panza persona attribution, across three trials.}
    \centering
    \resizebox{\linewidth}{!}{
    \begin{tabular}{lcc|aa|cc|aa}
        \toprule
        Trial No. & \multicolumn{2}{c}{Subject 1} & \multicolumn{2}{a}{Subject 2} & 
 \multicolumn{2}{c}{Jeff} &
 \multicolumn{2}{a}{Llama} \\
\cmidrule(lr){2-3}\cmidrule(lr){4-5} \cmidrule(lr){6-7} \cmidrule(lr){8-9}
 & \makecell{P} & \makecell{R}  & \makecell{P}  & \makecell{R} & \makecell{P} & \makecell{R} & \makecell{P} & \makecell{R}  \\
         \midrule
         1& 0.76  & 0.77 & 0.79 &  0.77 & 1.00 & 1.00 &  0.97 & 0.98   \\
         2    & 0.62  & 0.62 & 0.66 & 0.66 & 0.71 & 0.71 & 0.71 & 0.71  \\
         3   & 0.54 & 0.54  & 0.45 & 0.45 &0.67 & 0.67 & 0.94 & 0.94 \\
         \midrule
         Avg. & \multicolumn{4}{c|}{0.64 \quad 0.64} & 0.78 & 0.78 & \multicolumn{2}{c}{0.87 \quad 0.87}  \\
         \bottomrule
    \end{tabular}}

    \label{tab:persona_recognition}
\end{table}

\begin{table}[t]
    \caption{Number of authentic and LLM-generated emails found suspicious by acquaintances of the Panza training set authors.}
    \centering
    \resizebox{\linewidth}{!}{
    \begin{tabular}{c|a|c|a|c|a|c}
        \toprule
         Author & Subject 1 & Subject 2 & Subject 3 & Subject 4 & Subject 5 &Avg. (\%)  \\
         \midrule
         Genuine                & 1/5 & 2/5 & 0/5 & 1/5 & 2/5 & 24 \% \\
         \texttt{Llama-3-8B}    & 4/5 & 3/5 & 4/5 & 3/5 & 2/5 & 64 \% \\
         \texttt{Panza-L}   & 5/5 & 1/5 & 0/5 & 0/5 & 1/5 & 28 \% \\
         \texttt{Panza-S}   & 3/5 & 2/5 & 1/5 & 0/5 & 2/5 & 32 \% \\
         
         \bottomrule
    \end{tabular}}

    \label{tab:impersonation}
\end{table}

\subsection{Impersonation Capabilities}
\label{sec:impersonation_human_eval}
The final user study evaluates whether Panza can produce emails that have \textit{credibility} - in that a human familiar with someone's style would find the emails to be credibly written by that author. For this study, we recruited three pairs of volunteers, one of which served as the author and the other served as the evaluator; we required the evaluator to be personally acquainted with the author and their writing style. (Two of the pairs consisted of the same two people, in alternating roles). Each of the authors then finetuned two models from Llama3-8B using RoSA on their private data: a model trained on 80\% of their training data -- \texttt{Panza-L} -- and a model trained on a random sample of 75 emails from their training data -- \texttt{Panza-S}; the participants used the standard training hyperparameters for \texttt{Panza-L} and trained for a total of 11 epochs for \texttt{Panza-S}, matching the settings in Section~\ref{sec:ablation}. 

Then, for each user, we sampled a selection of $20$ emails from their held-out test set, and evenly distributed them into four groups of five; for each group, we selected one of the following responses:  i) the actual email written by the participant, ii) the output of a \texttt{Panza-L} model, iii) the output of a \texttt{Panza-S} model, and iii) the output of Llama-3-8B-Instruct model. These responses (without the prompts) were then sent to the evaluator associated with the author, who was then told that some of the emails were computer-generated, and asked to rate every response on a binary scale of whether the email is: i) suspicious -- likely written by a bot, or, ii) credible: the email sounds like something the author would have written. The evaluators were not told what proportion of emails were genuine, or which specific models were used to generate the responses in the study.

We present the results of this study in Table \ref{tab:impersonation}. We note that the experimental setup naturally makes the evaluators suspicious of the emails that they see, as they are asked explicitly to consider the genuineness of the emails they evaluate; something people likely don't do every time they see a new email. Even so, of the five emails shown to each user per category, on average between 28-32\% of the \texttt{Panza} emails were found suspicious - \emph{in other words, even in this setting, over two thirds of the emails generated by Panza were found credible}. By comparison, 24\% of the genuine emails were found suspicious, as were 64\% of the emails generated by the baseline Llama-3-8B model. 

These results show that, while emails that the participant wrote themselves were found the most credible on average, emails produced by Panza were also frequently viewed as credible; this happened far less frequently with a plain Llama model. This supports our conclusion that using a Panza model makes it feasible to create messages that are perceived as written by the original author, even when individuals are inclined to be suspicious of authorship.

\section{Limitations}

The techniques used by Panza provide a significant improvement, across a mix of metrics, with respect to on-device personalization, and, we believe, a compelling case of practically useful LLM personalization. Yet, more effort should be dedicated to 1) designing metrics for such novel personalization tasks, and also, more broadly 2) accurately measuring LLM performance on open-ended tasks such as text generation, with regard to improving its representation of the personal data of the user. Additionally, as a proof-of-concept, Panza has currently only been tested in English and on email data, leaving open the creation of such tools for other languages and types of personalized text generation. Our techniques should be easily extensible to such scenarios.

\section*{Impact statement}

We foresee two categories of risks from presenting a project like Panza. First, a malicious user who has access to a sufficient number of third party's emails (for instance, the ones received from that party, or available online through open mailing lists) can use a tool like Panza to create a credible malicious imitation of that individual. Second, a style-matching tool like Panza can be used to create derivative work that can credibly be misrepresented as original (for instance, for completing school assignments). Panza's low price and accessibility may aid in such misuse; however, we note that, the existence of open-source text generation models and finetuning methods, as well as the text personalization tools available through service-based models such as Google Gemini already give more sophisticated users access to these technologies.
We believe that the relatively high fidelity of emails generated using our approach is a strong incentive towards consistently implementing and using cryptographic techniques such as digital signatures in order to curtail the impact of such advanced personalized text-writing tools. 

\section*{Acknowledgments}
The authors would like to thank Michael Goin and Tony Wang for their feedback on this work, and their help in distributing the project. This research was supported by the Scientific Service Units (SSU) of IST Austria through resources provided by Scientific Computing (SciComp). EI was supported in part by the FWF DK VGSCO, grant agreement number W1260-N35.

\bibliography{custom_iclr}

\vspace{1.0cm}
\clearpage
\appendix

\onecolumn

\section{Panza under \texttt{15GB} GPU memory}
\label{appendix:panza-quant}
This setting is  interesting since it allows training and deploying Panza on a free Colab instance (NVIDIA T4 GPU). In addition to the GPU memory constraint, we also assume access to only \texttt{12GB} of CPU memory, as is the case for Colab. To accomplish this, we quantize multiple pipeline components to $4$ bits/parameter. 

\begin{enumerate}
    \item \textbf{Quantized base weights}. We store the weights of the base model in doubly-quantized $4$ bit precision \citep{dettmers2023qlora}, reducing the model size in memory by roughly $4\times$.
    \item \textbf{Quantized accumulators}. As part of the sparse adapter's mask generation, RoSA accumulates gradients of the base weights on CPU, violating the \texttt{12GB} CPU memory restriction. To remedy this issue, we change the precision of the accumulators to $4$ bits using a uniform quantization with group size $128$. 
    \item \textbf{Adapter merging}. To merge the half-precision RoSA adapters and the $4$-bit base weights, we present a new accurate merging mechanism; for each layer, we dequantize the base weights, add adapters, and quantize the result back to $4$ bits using the GPTQ method \citep{frantar2022gptq}. Our key contribution is an \emph{adapter-aware} implemetation of the GPTQ algorithm, where the quantization is interleaved with merging per layer, without materializing the full half-precision model.
\end{enumerate}

\subsection{The Impact of Compression}
We evaluate this memory-efficient version of Panza, which requires less than \texttt{15GB} of GPU memory, and \texttt{12GB} CPU RAM. We first examine how quantizing each component of the pipeline affects the results, and show that quantizing all the components at the same time can achieve reasonable results. We fine-tune the \texttt{Mistral-Instruct-7b-v0.2} model using RoSA on the \texttt{David} dataset, and use the same three BLEU, ROUGE, and MAUVE metrics for evaluation. 

\begin{table}[h!]
\centering
\caption{Effect of applying $4$-bit quantization to different components of Panza on the \texttt{David} dataset without RAG. We use \texttt{Mistral-Instruct-7b-v0.2} for both summarization and fine-tuning. \texttt{BNB} stands for \texttt{bitsandbytes} \citep{dettmers2023qlora} and GPTQ indicates the quantization technique of \citep{frantar2022gptq}. Further, \texttt{QSum}: summarization with quantized model, \texttt{QRoSA}: RoSA with quantized base model, \texttt{QAcc}: quantized gradient accumulation for RoSA, \texttt{QInf}: Inference with quantized model, and \texttt{QAll}: all components quantized at the same time.}
\label{tab:quant}
\scalebox{0.75}{
\begin{tabular}{lccc}
\toprule
                   & BLEU & ROUGE & MAUVE \\ \toprule
\texttt{All BF16} & $0.265$ & $0.432$ & $0.971$ \\ \toprule 
\texttt{QSum (BNB)} & $0.153$ & $0.283$ & $0.998$ \\
\texttt{QRoSA (BNB)} & $0.268$ & $0.417$ & $0.971$ \\
\texttt{QAcc (Uniform)} & $0.239$ & $0.407$ & $0.971$ \\
\texttt{QInf (BNB)} & $0.084$ & $0.192$ & $0.130$ \\
\texttt{QInf (GPTQ, ours)} & $0.251$ & $0.414$ & $0.971$ \\ \toprule
\texttt{QAll (BNB Inf)} & $0.076$ & $0.119$ & $0.070$ \\
\texttt{QAll (GPTQ Inf, ours)} & $0.207$ & $0.390$ & $0.996$ \\
\toprule
\end{tabular}
}
\end{table}

\paragraph{Quantizing components separately.} As described in Appendix \ref{appendix:panza-quant}, quantization can be alternatively applied to 1) RoSA base weights, 2) RoSA gradient accumulators, and 3) the final model. 
In addition, the summarization model can also be compressed. 
Table \ref{tab:quant} shows how quantizing each component affects the final accuracy. 
Results show that (a) quantizing most components individually only marginally affects the results (especially in terms of MAUVE score); (b) quantizing the inference model with \texttt{bitsandbytes} (\texttt{BNB}) \citep{dettmers2023qlora} significantly downgrades the accuracy, while (c) GPTQ quantization \citep{frantar2022gptq} recovers comparable accuracy to the original. Applying GPTQ to this setting is only possible due to our new implementation, described in Appendix \ref{appendix:panza-quant}, which allows for merging $4$-bit weights with half-precision  adapters without materializing a full model.

\paragraph{Fully-quantized Panza.} We apply quantization to every component at the same time and report the accuracy of the model in Table \ref{tab:quant}. These results show that the limitation of training using under \texttt{15GB} of GPU memory results in a $5.8$-point drop in BLEU score but unchanged ROUGE and MAUVE scores, suggesting a slightly worse but still useful model, relative to training in \texttt{BF16} precision.

\section{Panza as a Google Chrome Extension}\label{appendix:google_chrome_extension}
\begin{figure}[h!]
     \centering
     \includegraphics[width=0.85 \textwidth, height=5cm]{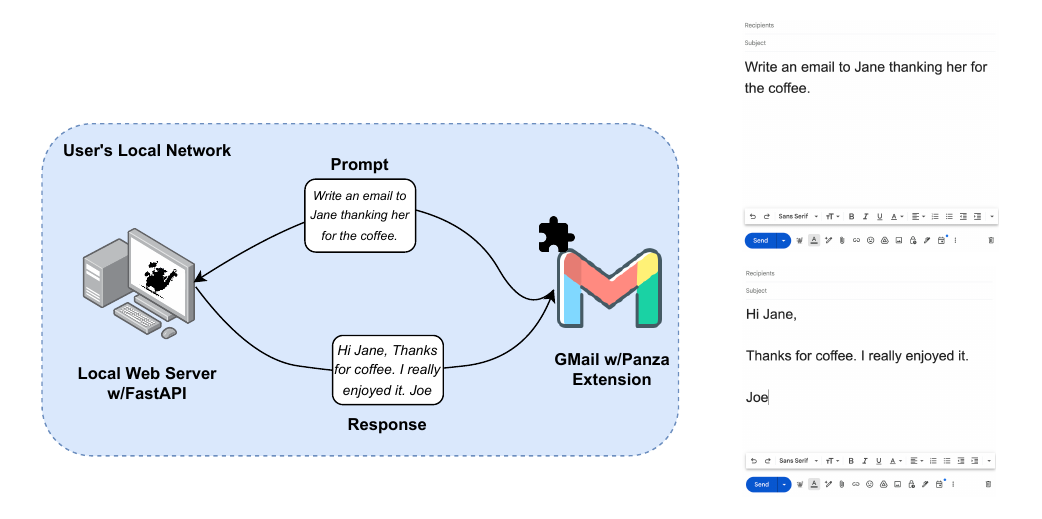}
     \caption{Panza delivered as a Google Chrome Extension}
     \label{fig:panza_ext_workflow}
\end{figure}

In addition to the entire personalization framework that Panza provides via finetuning, we also focus on how the end user will make use of their personalized email agent. Specifically, we want to integrate Panza directly into the user's GMail experience such that they can use the agent directly within their typical email workflow. We address this by creating a simple, \textit{local}, FastAPI application that is used to serve the user's Panza model. This is complimented by a \textit{publicly available} Google Chrome Extension which will to coordinate the interaction with the email agent from the user's perspective. To use Panza within GMail, the user first describes the prompt that they wish to send to the Panza model in the typical message dialog. Then, upon clicking the Panza button in the menu bar, the Google Chrome Extension sends a request to the FastAPI service, which in turn prompts the underlying model to generate an email response. This response send back to the Google Chrome Extension which is then used to populate the email draft in the place of the prompt that had been written by the user. The workflow is illustrated in Figure \ref{fig:panza_ext_workflow}.  

\section{Base model comparison}
\label{appendix:model_choice}
\begin{figure*}[h!]
  \centering
  \renewcommand{\arraystretch}{1}
  \includegraphics[height=.3\linewidth]{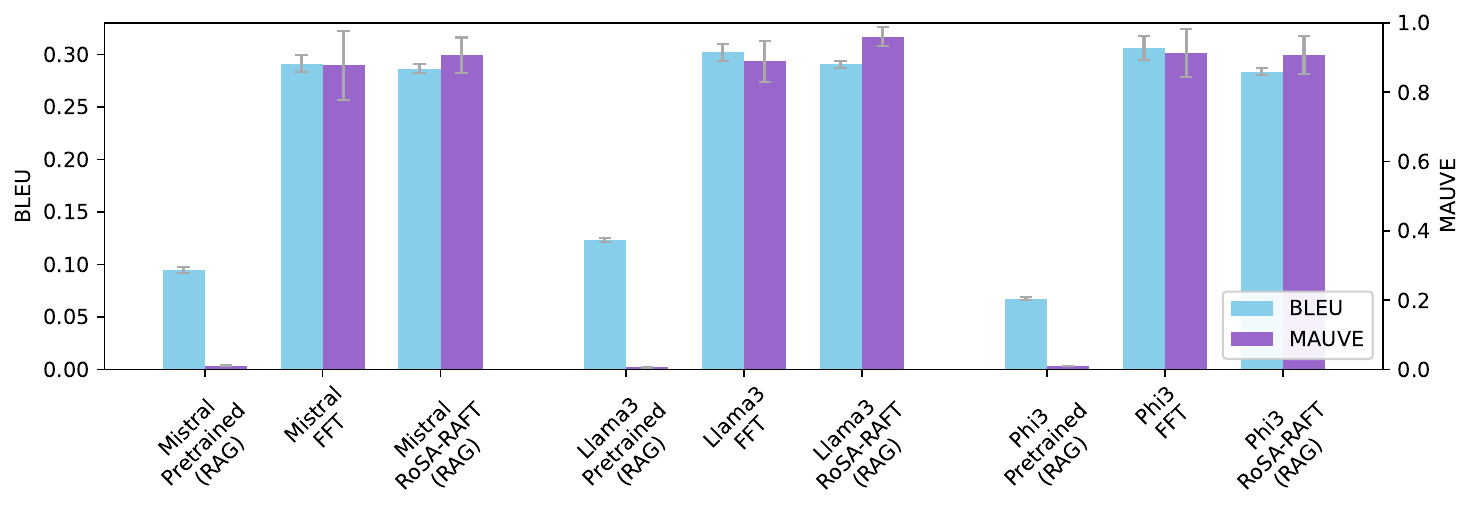}
  \vspace{-20pt}
  \caption{\label{figModels}
  Comparison between models for user Anon1. All models -- \texttt{Mistral-7B-Instruct-v0.2} (Mistral), \texttt{Meta-Llama-3-8B-Instruct} (Llama3), and \texttt{Phi-3-\-mini-\-4k-\-instruct} (Phi3) -- incorporate the user's style after fine-tuning and outperform the the Pretrained + RAG baseline. While BLEU scores are the same across models, MAUVE scores are higher on the Llama3 model.  
  }
\end{figure*}

We validate the usefulness of several popular open-source foundation models the base model for Panza, as well as our statement that fine-tuning, rather than simply RAG, is necessary to obtain a performant Panza model. In Figure~\ref{figModels} we show that similar performance levels can be obtained by fine-tuning various LLM backbones, when performing FFT or RoSA-RAFT across Mistral-7B-Instruct-v0.2, Llama3-8B-Instruct, and Phi-3-mini-4k-instruct. While the achieved BLEU scores are very similar across models, the only significant difference is the higher MAUVE score achieved by the Llama3 model.

\section{Prompt Engineering}
\label{sec:appendixPrompting}

In the first phase of finetuning, we generate summaries with the help of the following \textbf{summarization prompt}:
\begin{boxRound}
\begin{quote}
    """Summarize the following email that I wrote, in an imperative form, in one or two or maximum three sentences, and make sure to include relevant information, without copying the email content itself. The summary should look like an instruction directing someone to write the same email, and start with Instruction:\newline
    Here is the email text:\newline
    \{email\}""" 

\end{quote}
\end{boxRound}

Then, to generate emails, we give the instructions back to the model using the following format:
\begin{boxRound}
    \begin{quote}
    """\newline
    \{system preamble\}\newline
    \newline
    \{user preamble\}\newline
    \newline
    \{rag prompt\}\quad\# [optional] \newline
    \newline
    Instruction: \{instruction\}\newline"""
\end{quote}
\end{boxRound}

The \textbf{system preamble} sets the role of the LLM as follows:
\begin{boxRound}    
\begin{quote}
    """Your role is that of a helpful automated email assistant. I will provide you with a short instruction, and you have to write a well-formed email in my style following this instruction. Be sure to follow my email writing style!  In case you see a nonsensical instruction, you should not reply with an email, but with the expression "Sorry, but I don't get it."
    """
\end{quote}
\end{boxRound}

The \textbf{user preamble} provides optional information about the user. For the five users in our experiments, we set it to ``My name is \texttt{FirstName LastName}''. Generally, it can be filled with any relevant information about the user, for instance:
\begin{boxRound}
\begin{quote}
    """My name is Jane Doe. I work as a manager at Acme Corp. My address is 123 Main Street, Springfield, IL, USA. My boss's name is Alex Burns. My children's names are Elsa, Anna, and Olaf. I am deeply committed to my hobby of underwater basket weaving, for which we meet every Thursday at noon."""
\end{quote}
\end{boxRound}

Finally, for RAG we retrieve several relevant emails for the current instruction and include the following \textbf{rag prompt} to the input:
\begin{boxRound}
\begin{quote}
    """Extract specific information from these previous e-mails only if it is relevant to the current e-mail you have to write.\\
    \\
    Previous e-mails:\\
    \\
    EMAIL CONTENT:\\
    <email\_1 content>\\
    \\
    $---$\\
    \\
    EMAIL CONTENT:\\
    <email\_2 content>\\
    \\
    $---$\\
    $...$"""
\end{quote}
\end{boxRound}

\section{Summarization}
\label{sec:appendixSummarization}

\begin{table*}[t]
\centering
\caption{Summarization quality relative to ``golden'' user-generated instructions, across different models (10 seeds).}
\label{tab:summarization_metrics}
\scalebox{0.75}{

\begin{tabular}{l>{\columncolor{gray!10}}c>{\columncolor{gray!10}}ccc}%
\hline%
 &\multicolumn{2}{c}{\texttt{David}}&\multicolumn{2}{c}{\texttt{Anon1}}\\%
\hline%
Method&BLEU&ROUGE&BLEU&ROUGE\\%
\texttt{Mistral{-}7B{-}Instruct{-}v0.2}&$0.219 \pm 0.009$&$0.353 \pm 0.007$&$0.133 \pm 0.009$&$0.251 \pm 0.012$\\%
\texttt{Mistral{-}7B{-}Instruct{-}v0.3}&$0.229 \pm 0.021$&$0.373 \pm 0.021$&$0.164 \pm 0.014$&$0.307 \pm 0.016$\\%
\texttt{Phi{-}3{-}mini{-}4k{-}instruct}&$0.268 \pm 0.020$&$0.422 \pm 0.018$&$0.180 \pm 0.006$&$0.331 \pm 0.011$\\%
\texttt{Meta{-}Llama{-}3{-}8B{-}Instruct}&$\textbf{0.394} \pm 0.011$&$\textbf{0.548} \pm 0.013$&$\textbf{0.264} \pm 0.011$&$\textbf{0.411} \pm 0.015$\\%
\hline%
\end{tabular}%

}
\end{table*}

In Table~\ref{tab:summarization_metrics} we measure the summarization quality, by comparing against golden summaries from the users David and Anon2. In Table~\ref{tab:summarization_examples} we inspect several generated summaries for each model. The golden summaries used in this study were performed by the data donors, without compensation.

\begin{table*}[h]
\centering
\caption{Summarization comparison between ``golden'' user-generated instructions and summaries generated by different pre-trained LLMs.}
\label{tab:summarization_examples}
\scalebox{0.7}{

\begin{adjustbox}{center}

\begin{tabular}{|m{4.5cm}|m{4cm}|m{4cm}|m{4cm}|m{4.5cm}|}
\hline
\textbf{Original email} & \textbf{Golden summary} & \textbf{Llama-3-8B-Instruct} & \textbf{Mistral-7B-Instruct} & \textbf{Phi-3-mini-4k-instruct} \\ \hline
Hi Cheryl, The proposal looks good, maybe we can discuss it in a bit more detail tomorrow. Best regards, David & Write to Cheryl saying that the proposal looks good and suggest a meeting the following day. & Write an email to Cheryl, stating that the proposal looks good and suggesting a more detailed discussion for the next day. & Write an email to Cheryl suggesting a meeting to discuss the proposal in more detail tomorrow. End with a polite closing. & Write an email to Cheryl, acknowledging the good proposal, suggesting a detailed discussion for the following day, and signing off with your name. \\ \hline
Hi both, I'm fairly available in the afternoons next week, except for Monday and Friday, so please feel free to suggest a slot. Best, David & Write an email to two recipients, stating that I am available in the afternoons next week, except for Monday and Friday, and ask them to suggest a time that works for them. & Write an email to both parties, stating that you are available in the afternoons next week, excluding Monday and Friday, and inviting them to suggest a suitable time slot. & Write an email informing the recipients that you are generally available in the afternoons from Tuesday to Thursday next week, except for Mondays and Fridays. Request them to suggest a suitable time. & Write an email informing the recipients that you are generally available in the afternoons next week, except for Monday and Friday, and invite them to propose a suitable time slot for a meeting. \\ \hline
Dear Ernie, Thank you for the invitation. I would have loved to join, but I'm trying to save some time and energy for my other admin tasks this semester :-) Best regards, David & Write an email to Ernie, thanking him for the invitation, but saying that I have to decline due to a heavy admin burden this semester. & Write an email to Ernie declining an invitation, citing a need to prioritize other administrative tasks this semester. & Inform Ernie that you're unable to attend due to prior commitments with administrative tasks, expressing gratitude for the invitation. & Write an email to Ernie, expressing gratitude for an invitation, politely declining due to prior commitments related to admin tasks, and sign off with a warm closing. \\ \hline
Dear David, I was wondering if I could ask you for an "emergency-mode" code review (see attached). Specifically, it would be great if you could have a look at it in the next couple of weeks. I'd really appreciate it if you can make it; please let me know. Cheers, David & Write an email to David requesting an "emergency-mode" code review, politely asking them if they can review it within the next couple of weeks. & Write an email to David requesting an "emergency-mode" code review, attaching the relevant document, and asking him to review it within the next couple of weeks. & Ask David for an urgent code review, requesting him to look at the attached document within the next couple of weeks and kindly express appreciation for his assistance. & Write an email to David requesting an emergency-mode code review, emphasizing the need for it to be completed within the next couple of weeks, and express appreciation for his potential assistance, while also asking for his confirmation of availability. \\ \hline
\end{tabular}

\end{adjustbox}

}
\end{table*}

\section{Panza Generated Emails}
\label{sec:appendixGenerations}

\begin{table*}[h]
\centering
\caption{Generation comparison for \texttt{Llama-3-8B-Instruct} model fine-tuned for different users.}
\label{tab:generation_examples}
\scalebox{0.7}{

\begin{adjustbox}{center}

\begin{tabular}{|m{2.5cm}|m{6.8cm}|m{2.7cm}|m{3.0cm}|m{3.0cm}|m{3.0cm}|}
\hline
\textbf{Instruction} & \textbf{Llama-3-8B-Instruct (Pretrained)} & \textbf{David} & \textbf{Kay} & \textbf{Sara} & \textbf{Tana} \\ \hline
Write an email to two recipients, stating that I am available in the afternoons next week, except for Monday and Friday, and ask them to suggest a time that works for them. & Dear [Recipient 1] and [Recipient 2], \newline I hope this email finds you well. I wanted to touch base with you regarding my availability for a meeting next week. I am free in the afternoons, from Tuesday to Thursday, and would be happy to schedule a meeting with you during this time. \newline Please note that I am not available on Monday and Friday, as I have other commitments. If you could let me know a time that works for you, I would greatly appreciate it. \newline Best regards, \newline [Your Name] & Hi, \newline I am available in the afternoons next week, except for Monday and Friday. Could you please suggest a time that works for you? \newline Best regards, \newline David & I'm available in the afternoons next week, except for Monday and Friday. How about Tuesday or Wednesday? Please let me know what works for you. \newline Thanks, \newline Kay & I am available in the afternoons next week, except for Monday and Friday. Can you both let me know when you are available and we'll set something up. & I am available in the afternoons next week, except Monday and Friday. What time works for you? \\ \hline
Write an email to the group proposing to resume meetings tomorrow at 11:30 AM, and also mention your availability in the office on Thursday afternoon and Friday. & Dear team, \newline I hope this email finds you well. As we've been on a brief hiatus, I'd like to propose that we resume our regular meetings tomorrow at 11:30 AM. I believe it's essential to catch up and discuss ongoing projects, and I'm confident that this will help us stay on track. \newline Additionally, I wanted to let you know that I'll be available in the office on Thursday afternoon and Friday. If any concerns need to meet or discuss any matters, please don't hesitate to reach out to me. \newline Looking forward to seeing you all tomorrow at 11:30 AM. \newline Best regards, \newline [Your Name] & "Hi all, \newline We should probably resume the meetings tomorrow at 11:30 AM. Also, I am available in the office on Thursday afternoon and Friday. \newline Best, \newline David" & I'm thinking we should resume meetings tomorrow at 11:30 AM. I'm in the office tomorrow afternoon, and I'll be here Thursday afternoon, and Friday. \newline I hope this helps. \newline Kay & I'm going to propose that we resume our meetings tomorrow at 11:30 AM. I'll be in the office tomorrow and Thursday afternoon. I'll be out of the office on Friday. Please let me know if you have any questions. \newline Thanks, \newline Sara & I think we should try to resume our meetings tomorrow at 11:30 AM. I'll be in the office tomorrow afternoon, Thursday afternoon and Friday. \\ \hline
Write an email to Jeannie, thanking her for the monitor, and then request assistance in gaining access to the conference room after hours, as your current ID does not provide access. & Dear Jeannie, \newline I wanted to take a moment to express my gratitude for the new monitor you provided for me. It has been a huge help in my work and I appreciate your support in keeping everything up to date. \newline On a separate note, I was wondering if you could assist me with gaining access to the conference room after hours. As you may know, my current ID does not provide access to the room outside of regular working hours. I often find myself needing to work late [...] & Hi Jeannie, \newline Thanks for the monitor! Unfortunately, my current ID does not provide access to the conference room after hours. Could you please help me with this? \newline Best regards, \newline David & Thanks very much for the monitor. I really appreciate it. I'm having some trouble getting into the conference room after hours. I've got a new ID, but it doesn't seem to work. Could you help me out? \newline Thanks a bunch. \newline Best, \newline Jeff & Jeannie: Thanks for the monitor. I was wondering if you could help me with something. I need to be able to access the conference room after hours and my current ID does not provide access. Can you help me with this? \newline Thanks, \newline Sara & Thanks for the monitor, Jeannie! It is a great help. Now, I need your help with one more thing. My new ID doesn't seem to give me access to the conference room after hours. Can you help me out with this? \newline Thanks! \\ \hline
\end{tabular}

\end{adjustbox}

}
\end{table*}

Table~\ref{tab:generation_examples} compares emails generated by models fine-tuned for different users, highlighting their style differences.

\section{Details of general knowledge measurement}
\label{sec:general_knowledge_extended}

All results are obtained through the corresponding \textit{lm-evaluation-harness} framework~\citep{eval-harness}, and are reported in Table~\ref{tab:general_knowledge}. 
Based on the results collected for \texttt{David}, we conclude that personalized Panza models retain, on average, over 98\% of the baseline model's accuracy. This suggests that the Panza personalization process avoids catastrophic forgetting.

\begin{table*}[h!]
\centering
\caption{Performance of personalized Panza models on general knowledge benchmarks.}
\label{tab:general_knowledge}
\small{
\setlength{\tabcolsep}{5pt}
\begin{adjustbox}{max width=\textwidth}
\begin{tabular}{lcccc}
\toprule
\textbf{} & \makecell{Llama-3-8B\\Instruct} & FFT &  LoRA &  RoSA  \\
\midrule
Arc-c & 60.7 & 60.3 &  60.7 &  58.9  \\
MMLU & 67.1 & 65.5 & 65.6 &  65.3  \\
Hellaswag & 78.5 & 78.7 & 78.6 & 78.2  \\
Winogrande & 74.5 & 74.3 & 74.9 & 73.2 \\
GSM8k & 68.7 & 70.2 & 69.1 & 68.4 \\
TruthfulQA & 51.6 & 50.9 & 51.4 & 50.7 \\
\hline
Average & 66.9 & 66.7 & 66.7 & 65.8  \\
Recovery (\%) & 100 & 99.7 & 99.7 & 98.4 \\
\bottomrule
\end{tabular}
\end{adjustbox}
}
\end{table*}

\section{Comparison to LongForm}
\label{sec:longform_comparison}

We compare Panza to the results of~\citet{koksal2023longform}, which introduces Reverse Instructions and tests this method on various forms of long-form data outputs, including emails from the Enron dataset. We note that this is not a fair comparison, as the LongForm models are trained from a non-instruction-tuned model, are finetuned on a dataset that is only about 2\% emails, and make no attempt at personalization (the emails come from multiple authors). Further, we note that the specific evaluation subset of the Enron emails differs in our evaluations, and it is not possible to bridge this gap (i.e., evaluate Panza on the LongForm Enron evaluation set), as Panza requires the availability of additional emails from the same author for creating a personalized model. We therefore present these numbers  only as an indication that the overall combination of instruction-tuned base model + Panza performs better than approaches that also use Enron emails and Reverse Instructions to create long-form content. Our results in Sections~\ref{sec:gpt} and \ref{appendix:model_choice}  demonstrate the superiority of Panza over base models + RAG.

\begin{table*}[h!]
\centering
\caption{Comparison of METEOR scores of Panza emails to LongForm~\citep{koksal2023longform}. All Panza models are trained using the RoSA-RAFT-RAG method with default hyperparameter values.}
\label{tab:meteor_longform}
\small{
\setlength{\tabcolsep}{5pt}
\begin{adjustbox}{max width=\textwidth}
\begin{tabular}{c|c|cccc}
\toprule
 LongForm & \multicolumn{5}{c}{Panza} \\
 \cmidrule(lr){1-1}\cmidrule(lr){2-6}
Enron Avg. & Avg. & Tana & Sara & Kay & Jeff \\

\midrule
8.9 & 36.3 & 40.4 & 38.6 & 33.8 & 31.3 \\
\bottomrule
\end{tabular}
\end{adjustbox}
}
\end{table*}

We present the results in Table~\ref{tab:meteor_longform}. The Panza-RAFT-RAG models achieve an average METEOR score of 36.3, while \citet{koksal2023longform} report a METEOR score of 8.9 on their best model on this task (LongForm-OPT-6.7B). We emphasize again that, due to methodological differences, this is not a fair comparison, but nevertheless believe that these results demonstrate the effectiveness of Panza on the task of email generation. 

\section{Comparison to GPT-4o}
\label{sec:gpt}
While the setting of uploading a user’s email history to a client such as ChatGPT does not meet our privacy requirements, it is still valid to ask whether a more capable LLM equipped with the user's entire email history (i.e., the training set) would perform as well as Panza. We tested this on users Anon1 and Anon2 by using the OpenAI GPT4o~\citep{openai2023gpt} API to generate emails after being prompted with the e-mail generation prompt and the user’s entire email history. The results are shown below for the two Anon users. For both, the BLEU score was higher when using Panza with RoSA, and the Mauve (personalization) scores of the OpenAI outputs were near 0, showing that the e-mail style was not personalized to the user. The Panza results are also presented below for easier comparison.

\begin{table*}[h!]
\centering
\caption{Comparison of BLEU and MAUVE scores of Panza emails to those generated by GPT-4o prompted with the user's entire email history. All Panza models are trained using the RoSA-RAFT-RAG method with default hyperparameter values.}
\label{tab:gpt}
\small{
\setlength{\tabcolsep}{5pt}
\begin{adjustbox}{max width=\textwidth}
\begin{tabular}{c|cc|cc}
\toprule
 & \multicolumn{2}{c}{BLEU} &  \multicolumn{2}{c}{MAUVE} \\
 \cmidrule(lr){2-3} \cmidrule(lr){4-5}
 User & GPT-4o & Panza & GPT-4o & Panza \\
 \midrule
 Anon1 & 0.24 & 0.31 & 0.009 & 0.96 \\
 Anon2 & 0.19 & 0.22 & 0.004 & 0.93 \\
\bottomrule
\end{tabular}
\end{adjustbox}
}
\end{table*}

\section{Alternate metrics: METEOR, BERT-score}
\label{appendix:alternate_metrics}

\begin{figure*}[h!]
  \centering
  \renewcommand{\arraystretch}{1}
  \includegraphics[height=.27\linewidth, width=\textwidth]{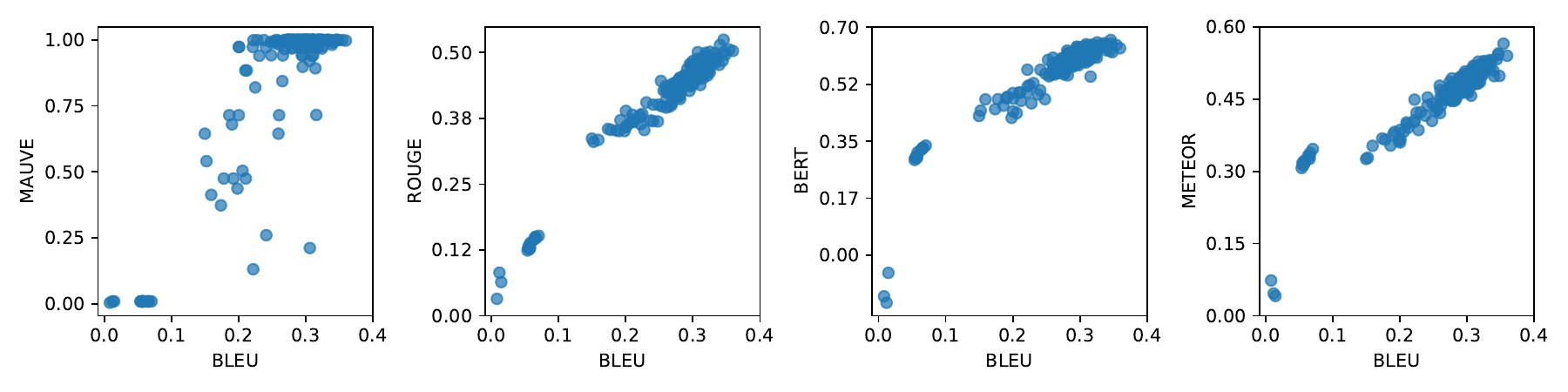}
  \vspace{-20pt}
  \caption{\label{metricsComparison}
  Similarity of BLEU to MAUVE, ROUGE, BERT-score, and METEOR for finetuned models. Note the near-perfect correlation between BLEU and ROUGE, BERT, and METEOR. By contrast, while BERT and MAUVE have some correlation, we observe models with low BERT scores and high MAUVE scores, and vice-versa. Models trained with RoSA, LoRA, and FFT are presented together; RAG is not used for inference in this evaluation.
  }
\end{figure*}

\begin{table*}[h!]
\centering
\caption{METEOR and BLEU scores across users and models. While METEOR scores for (un-finetuned) Llama models are similar to those for FFT and LoRA models for 4/5 users (Jeff, Tana, Kay, and Sara), we observe large differences for all users in the BLEU scores. Note that finetuned models (both RoSA and FFT) were preferred by users in our study. }
\label{tab:meteor}
\scalebox{0.75}{

\begin{tabular}{l|ccc|ccc}%
\hline%
 & \multicolumn{3}{c}{METEOR} &  \multicolumn{3}{c}{BLEU} \\
 \cmidrule(lr){2-4} \cmidrule(lr){5-7}
User & Llama+RAG & FFT-RAFT-RAG & LoRA-RAFT-RAG & Llama+RAG & FFT-RAFT-RAG & LoRA-RAFT-RAG \\
\midrule
Jeff  &  0.32 & 0.31 & 0.32 & 0.12 & 0.18 & 0.19 \\
Tana & 0.39 & 0.42 & 0.43 & 0.16 & 0.28 & 0.29 \\
Kay & 0.34 & 0.34 & 0.35 & 0.12 & 0.19 & 0.21 \\
Sara & 0.40 & 0.41 & 0.41 & 0.15 & 0.26 & 0.28 \\
David & 0.42 & 0.49 & 0.51 & 0.11 & 0.31 & 0.33 \\

\bottomrule
\end{tabular}%

}
\end{table*}

In addition to BLEU, ROUGE, and MAUVE, which we present in the main part of the paper, we considered two alternative metrics: BERT-score~\citep{zhang2020bertscore} and METEOR~\citep{Lavie2007METEOR}, using the Huggingface implementations for both with default hyperparameters. We use the \texttt{distilbert-base-uncased} model for computing the BERT embeddings.

BERT score is frequently used to evaluate text generation quality; unlike BLEU and ROUGE, it uses embeddings in a Large Language Model to compute similarity. While BERT would have also been a good choice to measure the phrasing quality, our experiments found it to be very highly correlated with the BLEU score, and therefore redundant, as we show in Figure~\ref{metricsComparison}.

The METEOR score is a variant of token-matching scores such as BLEU and ROUGE, but with several changes, notably using a weighted harmonic mean of the precision and recall, and allowing word synonyms. Like BERT, it is highly redundant with BLEU when used as a finetuning quality metric (see Figure~\ref{metricsComparison}). Notably, however, in our experiments, we found that METEOR scores for un-finetuned Llama+RAG models were very close to Panza models in several of the datasets considered. We demonstrate this effect in Table~\ref{tab:meteor}. Note the contrast to BLEU, which demonstrates clear differences in non-finetuned versus finetuned models, which also agrees with our results in the human study in Section~\ref{sec:usefulness}, which demonstrate that finetuned models outperform base models in the usefulness of their email generations.

We conjecture that this is due to METEOR allowing of synonyms when doing the token matching, which therefore does not take into account the specific word choices typical to someone's style, and thus prefer the BLEU score for our evaluation and hyperparameter tuning.

\section{Additional Human Studies - BLEU/MAUVE Validation}
\label{sec:additional_human_studies}
We conducted a human study to validate our hypothesis that models with high BLEU and MAUVE scores are better suited for the e-mail generation task than those with both of these scores low, or with one low and the other high.

\begin{figure}[h!]
  \centering
  \renewcommand{\arraystretch}{1}
  \includegraphics[height=.55\linewidth]{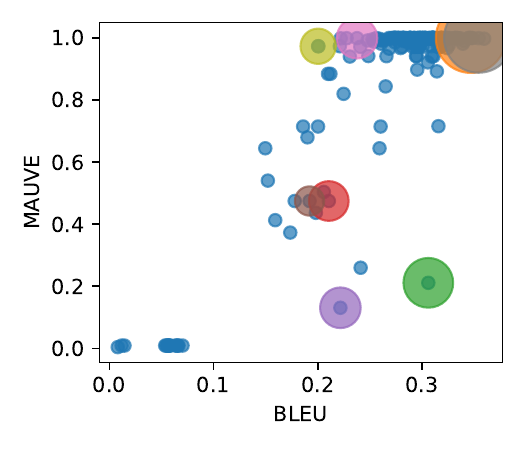}
  \caption{\label{david_bleu_mauve}
  Human-evaluated usefulness of emails output by various Panza models, plotted against their BLEU and MAUVE score. The size of the colored circles corresponds to the human-rated usefulness of the model; bigger signifies more useful.
  }
\end{figure}

As shown in Figure~\ref{metricsComparison}, during the course of hyperparameter tuning, we were able to find models that performed relatively well in BLEU, MAUVE, neither, and both. We selected eight of these models, roughly corresponding to each of these four cases (Both low, High BLEU/low MAUVE, low BLEU/high MAUVE, both high) for additional human evaluation. The donor of the dataset was asked to evaluate, on the same 3-point scale described in Appendix~\ref{appendix:usefulness_study}, with 3 being the highest score and 1 being the lowest. The results, superimposed onto the BLEU/MAUVE scores are shown in Figure~\ref{david_bleu_mauve}. We observe that the two models chosen for the highest BLEU and MAUVE scores are found to be the most useful, with average scores of 2.6. The model with high BLEU but lower MAUVE (green circle) was found to be less useful with a score of 2.125, indicating that generally the output emails required additional editing. Models with lower BLEU scores, or lower BLEU and MAUVE scores all had average ratings of 1.83 or below, with the lowest performing model (yellow circle) having a score of 1.36. We interpret these findings to suggest that, to maximize performance, both a (relatively) high BLEU and a high ROUGE score is necessary, and achievable.

This human study took the author of the David email dataset around 30 minutes, and the author was not compensated for this time.

\section{Human Study Details}\label{appendix:full_user_study_details}
\paragraph{Subject recruitment} In all user studies, due to the sensitive nature of the data, subjects were selected for their technical ability to conduct all steps of the evaluation - model training, evaluation data extraction, and the evaluation - without direct assistance from the experimenters, and therefore without exposing the experimenters to the subjects' private data. Thus, it was not possible to use any evaluation services for this task, and instead six subjects were recruited directly as volunteers. All subjects hold technical degrees, and were therefore well-suited for the task. The active time required to complete the tasks was 1-2 hours for each subject, ignoring passive tasks such as waiting for the email data export, and model training. Furthermore, each participant in the user study had between 199-979 emails in their training dataset; this was to ensure that the models had enough signal during the fine-tuning procedure to produce effective personalization, without overfitiing to the specifics of a few email samples. We note that the human rating tasks, after generating the necessary data for each task, took the subjects 30 minutes each, and the subjects were not compensated for their involvement in the user study.
\subsection{Additional Details for the Usefulness Study}
\label{appendix:usefulness_study}

As Panza is intended to convert relative prompts into long-form text, scores such as BLEU, and METEOR, that seek to match Panza-generated responses to those created by actual humans misses the nuance, that there is no one correct way to write a given email, and so we do not expect anything close to a perfect match between the Panza output and the email actually written by the user. Therefore, we conducted a human evaluation to measure the direct usefulness of the Panza models. In this evaluation, two subjects were asked to train Panza models on their own e-mail data, then evaluate the outputs of this model, as well as the outputs of Llama3-8B-Instruct augmented with RAG. Below, we detail the study and provide results from the evaluation in more detail.

\paragraph{Data preparation} The subjects were instructed to export emails from their provider of choice, with the loose guidance that it was not necessary to obtain more than 1000 emails. The subjects then ran a script that split the data to create a held-out validation set, then finetuned Panza-FFT-RAG and Panza-RoSA-RAG models from a \texttt{Llama3-8B-Instruct} checkpoint, using default hyperparameters and no additional tuning. Once the models were prepared, the subjects ran a provided script that, in the first experiment, generated Panza responses (emails) for 16 hand-constructed prompts shown in Table~\ref{tab:usefulness_eval}, and, in the second experiment, randomly selected 16 prompts from the user's held-out set and likewise generated Panza responses for these emails. Users were instructed to run every script once, and only the first generation (single attempt) was allowed for each prompt per model. In addition to Panza outputs, the subjects also extracted the same responses for an un-finetuned Llama3-8B-Instruct checkpoint augmented with RAG. In all cases for each user, the RAG database was the same, and did not contain the held-out emails.

The resulting prompts and responses(emails) were output in the form of a \texttt{.csv} file, and the subjects were instructed to import the file into Google sheets and conduct the evaluation there. Some guidance was provided on ways to format the spreadsheet for easy reading.

\paragraph{Rating instructions} The subjects were asked to rate the output emails on a 3-point scale. Specifically, they were given the following instructions:

Please rate the email outputs. The rating scale is as follows. Please use your own best judgment for what exactly minor/nontrivial changes mean to you:
\begin{itemize}
    \item 3: Very useful: would send as-is or with minor changes
    \item 2: Moderately useful: would sent only after nontrivial changes
    \item 1: Not useful: needs a complete rewrite
\end{itemize}

The ratings were then averaged for the final score. We present the 16 hand-constructed prompts, as well as the ratings each subject gave their models on this task, in Table~\ref{tab:usefulness_eval}. For privacy reasons, we are not able to provide the Panza outputs for these prompts, nor any of the User-specific prompts or their outputs.

\begin{table*}[h]
\centering
\caption{Panza performance on prompts from the fixed-prompt dataset.}
\label{tab:fixed_prompts}
\resizebox{\textwidth}{!}{
\begin{tabular}{|m{10.0cm}|a|a|a|c|c|c|a|a|a|c|}
\toprule
& \multicolumn{3}{a}{Subject 1}& \multicolumn{3}{c}{Subject 2} & \multicolumn{3}{a}{Subject 3} & \\
\cmidrule(lr){2-4} \cmidrule(lr){5-7} \cmidrule(lr){8-11}
\textbf{Prompt} & \textbf{Llama} & \textbf{FFT} & \textbf{RoSA} & \textbf{Llama} & \textbf{FFT} & \textbf{RoSA} & \textbf{Llama} & \textbf{FFT} & \textbf{RoSA} & Average \\
\midrule
Write an email to tell my two good friends Ina and George \newline that I won't be able to make it to dinner on friday. Make sure to add an apology and say I hope to see them soon. & 2 & 3 & 2 & 3 & 1 & 1 & 3 & 2 & 1 & 2.00\\
\midrule
Write an email inviting a large group of people to dinner at my house. Tell them I'm excited to try out my new recipe for roasted fish. & 2 & 2 & 3 & 1 & 1 & 1 & 2 & 2 & 3 & 1.89\\
\midrule
Write an email to Ziqui Chen telling her that I am still interested in buying the microwave from her, and I could pick it up next Thursday or Friday. & 3 & 3 & 3 & 3 & 3 & 3 & 3 & 2 & 3 & 2.89\\
\midrule
Write an email to a travel agency, telling them that you and your parents would prefer Australia over Myanmar, and would like to have modern accomodations with good showers. & 2 & 3 & 3 & 2 & 1 & 1 & 1 & 2 & 3 & 2.00\\
\midrule
Write an email to my good friend Alice asking her if she has any recommendations for what to do in Peru, when I travel there next summer. & 1 & 3 & 2 & 1 & 3 & 3 & 3 & 3& 2 & 2.33\\
\midrule
Write an email to my former coworker, apoligizing for the late response and telling her that I'd be happy to grab lunch or coffee when she comes to Vienna for ICLR. & 2 & 1 & 2 & 2 & 3 & 3 & 2& 2 & 3& 2.22\\
\midrule
Write an email to James Chen to thank him for his interest in the group, but telling him that it's not clear if any interns can be hired this year. & 2 & 3 & 2 & 2 & 1 & 2 & 2& 1 & 3& 2.00\\
\midrule
Write an email to Anna Karina to tell her that you'd be happy to answer any questions she has about the group, but you will likely be slow to reply for the next few weeks, due to upcoming conference submission deadlines. & 3 & 3 & 3 & 2 & 3 & 3 & 3& 3 & 3& 2.89\\
\midrule
Write an email to the immigration authorities to tell them that I urgently need an appointment to replace my lost residence permit, but unfortunately it seems that there is no availability on their website for the last month. & 3 & 3 & 2 & 2 & 1 & 2 & 1& 1 & 2& 1.89\\
\midrule
Write an email to the university administration to complain that the shuttle bus has been over 10 minutes late every day for the last couple of weeks. & 2 & 2 & 3 & 2 & 2 & 3 & 2& 1 & 3 & 2.22\\
\midrule
Write an email to former group member Alex, telling him that we can't find his laptop and asking where he left it. & 2 & 3 & 3 & 3 & 3 & 3 &2 & 1& 2 & 2.44\\
\midrule
Write an email to Katie showing her how to write a bash script that iterates over a directory and appends a string to every file. & 2 & 2 & 2 & 3 & 2 & 3 & 3& 1 & 2 & 2.22\\
\midrule
Write an email to my friend Albert with a recipe for banana bread. & 2 & 3 & 3 & 2 & 1 & 1 &2 & 1 & 3 & 2.00\\
\midrule
Write an email to my trainee Eric explaining the review process at NeurIPS and how to be a good reviewer. & 2 & 2 & 2 & 3 & 3 & 1 &3 & 2 & 3 & 2.33\\
\midrule
Write an email to Peter to tell him my new address. & 2 & 3 & 2 & 2 & 2 & 2 & 3& 3 & 3 & 2.44\\
\midrule
Write an email to Leandra to ask her to call me this evening at my phone number. & 3 & 3 & 3 & 3 & 2 & 2 & 3& 2& 2 & 2.56\\
\midrule
Average & 2.19 & 2.63 & 2.50 & 2.25 & 2.00 & 2.13 & 2.38 & 1.81 & 2.56 & 2.27\\

\bottomrule
\end{tabular}
}
\end{table*}

\begin{table*}[h]
\centering
\caption{Panza performance on prompts from the fixed-prompt dataset.}
\label{tab:fixed_prompts_2}
\resizebox{\textwidth}{!}{
\begin{tabular}{|m{10.0cm}|a|a|a|c|c|c|a|a|a|c|}
\toprule
& \multicolumn{3}{a}{Subject 4}& \multicolumn{3}{c}{Subject 5} & \multicolumn{3}{a}{Subject 6} & \\
\cmidrule(lr){2-4} \cmidrule(lr){5-7} \cmidrule(lr){8-10}
\textbf{Prompt} & \textbf{Llama} & \textbf{FFT} & \textbf{RoSA} & \textbf{Llama} & \textbf{FFT} & \textbf{RoSA} & \textbf{Llama} & \textbf{FFT} & \textbf{RoSA} & Average \\
\midrule
Write an email to tell my two good friends Ina and George \newline that I won't be able to make it to dinner on friday. Make sure to add an apology and say I hope to see them soon. & 1 & 3 & 3 & 2 & 3 & 2 & 3 & 3 & 3 & 2.56 \\
\midrule
Write an email inviting a large group of people to dinner at my house. Tell them I'm excited to try out my new recipe for roasted fish. & 1 & 2 & 3 & 1 & 3 & 3 & 2 & 2 & 2 & 2.11\\
\midrule
Write an email to Ziqui Chen telling her that I am still interested in buying the microwave from her, and I could pick it up next Thursday or Friday. & 1 & 3 & 3 & 3 & 3 & 3 & 3 & 3 & 3 & 2.78\\
\midrule
Write an email to a travel agency, telling them that you and your parents would prefer Australia over Myanmar, and would like to have modern accomodations with good showers. & 1 & 3 & 3 & 1 & 3 & 3 & 3 & 3 & 3 & 2.56\\
\midrule
Write an email to my good friend Alice asking her if she has any recommendations for what to do in Peru, when I travel there next summer. & 1 & 3 & 3 & 1 & 3 & 3 & 3 & 3 & 3 & 2.56 \\
\midrule
Write an email to my former coworker, apoligizing for the late response and telling her that I'd be happy to grab lunch or coffee when she comes to Vienna for ICLR. & 1 & 3 & 3 & 2 & 3 & 3 & 3 & 2 & 3 & 2.56\\
\midrule
Write an email to James Chen to thank him for his interest in the group, but telling him that it's not clear if any interns can be hired this year. & 1 & 3 & 3 & 2 & 2 & 3 & 3 & 3 & 3 & 2.56\\
\midrule
Write an email to Anna Karina to tell her that you'd be happy to answer any questions she has about the group, but you will likely be slow to reply for the next few weeks, due to upcoming conference submission deadlines. & 2 & 3 & 3 & 3 & 3 & 3 & 3 & 3 & 3 & 2.89\\
\midrule
Write an email to the immigration authorities to tell them that I urgently need an appointment to replace my lost residence permit, but unfortunately it seems that there is no availability on their website for the last month. & 1 & 3 & 3 & 2 & 2 & 3 & 3 & 3 & 3 & 2.56\\
\midrule
Write an email to the university administration to complain that the shuttle bus has been over 10 minutes late every day for the last couple of weeks. & 1 & 3 & 2 & 2 & 3 & 3 & 3 & 1 & 1 & 2.00\\
\midrule
Write an email to former group member Alex, telling him that we can't find his laptop and asking where he left it. & 1 & 2 & 3 & 1 & 2 & 2 & 3 & 1 & 2 & 1.89\\
\midrule
Write an email to Katie showing her how to write a bash script that iterates over a directory and appends a string to every file. & 1 & 3 & 3 & 1 & 2 & 3 & 3 & 2 & 2 & 2.22\\
\midrule
Write an email to my friend Albert with a recipe for banana bread. & 2 & 3 & 3 & 2 & 3 & 3 &3 & 1 & 3 & 2.56\\
\midrule
Write an email to my trainee Eric explaining the review process at NeurIPS and how to be a good reviewer. & 2 & 3 & 3 & 1 & 3 & 3 & 3 & 2 & 1 & 2.33\\
\midrule
Write an email to Peter to tell him my new address. & 3 & 3 & 3 & 2 & 3 & 3 & 3 & 2 & 3 & 2.78\\
\midrule
Write an email to Leandra to ask her to call me this evening at my phone number. & 2 & 3 & 3 & 1 & 2 & 3 & 3 & 1 & 2 & 2.22\\
\midrule
Average & 1.38 & 2.88 & 2.94 & 1.69 & 2.69 & 2.88 & 2.94 & 2.19 & 2.50 & 2.45\\

\bottomrule
\end{tabular}
}
\end{table*}

\subsection{Additional Details for the Persona Recognition Study}
Following the setup that was conducted for the usefulness study in Appendix \ref{appendix:usefulness_study}, each user was asked to perform inference on a further set of prompts to generate the responses necessary for the recognition study. In particular, each participant used their Panza-RoSA-RAG model and performed inference on a set of 24 \textit{custom prompts}, which were selected by the organising team\footnote{It is worthwile to note that these prompts were written such that they would not elicit any sensitive information from the user.}. As before, users were instructed to run this script once, and only the first generation (single attempt) was allowed for each prompt per model. The resulting prompts and responses(emails) were output in the form of a \texttt{.csv} file, and the subjects were instructed share this file with the experimenting team \textit{without observing} the contents of the file. This was done to protect the integrity of the study as, as will be come clear in the next paragraph, users will be asked to associate emails with users.

\begin{figure}
    \centering
    \resizebox{\textwidth}{!}{
    \begin{minipage}[h]{0.3\textwidth}
        \centering
        \includegraphics[width=\textwidth]{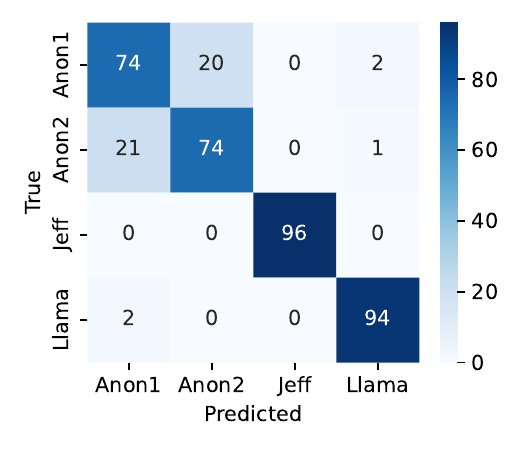}
    \end{minipage}
    \hfill
    \begin{minipage}[h]{0.3\textwidth}
        \centering
        \includegraphics[width=\textwidth]{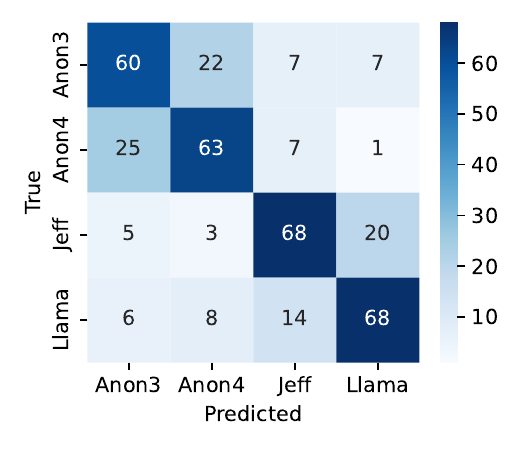}
    \end{minipage}
    \hfill
    \begin{minipage}[h]{0.3\textwidth}
        \centering
        \includegraphics[width=\textwidth]{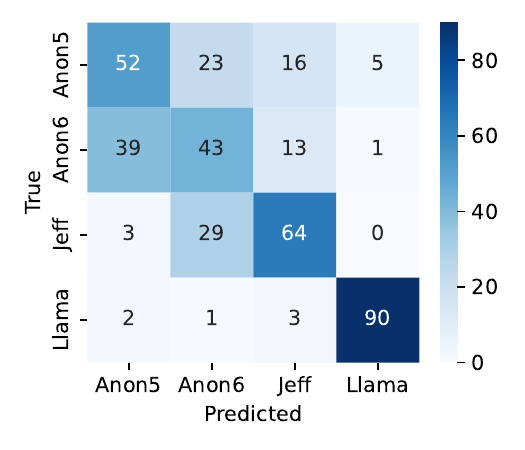}
    \end{minipage}}
    \caption{Confusion matrices for three independent experiments of the persona recognition study.}
    \label{fig:persona_recog_cm}
    
\end{figure}

Once the organising team received the outputs of from each of the four participants in a group, first each of the responses were cleaned and anonymised. To ensure that no personally identifiable information was revealed that would compromise the point of the recognition task (i.e the person's name). Next, the order of responses for each prompt were shuffled randomly, and outputted to a `.csv` file that was sent to each of the four participants for the rating task. In particular, each participant was asked to, for each of the 24 prompts, to assign each response to one of the four possible individuals involved in their study. We show the three confusion matrices for each independent experiment we conducted in Figure \ref{fig:persona_recog_cm}.

\subsection{Additional Details for the Impersonation Study}
The purpose of the impersonation study is to assess whether Panza models can produce emails that are believably like those of the user in whose data it was trained. The reason behind comparing both \texttt{Panza-L} and \texttt{Panza-S} was was that given our findings in Section \ref{sec:ablation}, we wanted to go beyond the metrics that we had selected for this work and evaluate whether the \textit{more data efficient} model is, like the large model, able to convincingly imitate the human qualities of the authors text generation. The user's gold output and the outputs from the \texttt{Llama-3-8B} models were chosen to be controls in the experiment. Namely, our hypothesis is that all of the authors own responses should be believable, whilst all the outputs of the \texttt{Llama} model would display the converse.

In addition to training the \texttt{Panza-L} model, all users were instructed to their their \texttt{Panza-S} model using the RoSA finetuning method with the default hyperparameters as per Section \ref{sec:ablation}. As before, users were instructed to run the command for producing the outputs for the impersonation study only once, and only the first generation (single attempt) was allowed for each prompt per model. Finally, the corresponding prompts for each response were removed, so that the rater only received the responses themselves. This was to ensure that we simulated a setting where users receive and email, along with no other additional context that could be used to determine the credibility of the email. The resulting responses(emails) were output in the form of a \texttt{.csv} file, where all participants were given a chance to clean the the outputs before sending their model outputs to organising team, who would then coordinate the experiment with the respective partner.

\section{Fine-Tuning Performance}
\label{sec:appendixFinetuning}

We show complete results for all models, across every user in Tables~\ref{tab:results_llama3} and \ref{tab:results_llama3_anon} (Meta-Llama-3-8B-Instruct), Tables~\ref{tab:results_mistral_v2} and \ref{tab:results_mistral2_anon} (Mistral-7B-Instruct-v0.2), and Tables~\ref{tab:results_phi3} and \ref{tab:results_phi3_anon} (Phi-3-mini-4k-instruct). To compare between different models, we report the average results over all users in Table~\ref{tab:results_all_models}. We find that for all the models, finetuning successfully incorporates the user's style. In Figure~\ref{figConfusionMatrices} we perform a pairwise style comparison for model trained on different users and the test emails of all the other users. This shows MAUVE score successfully captures style differences, while BLEU/ROUGE scores are limited to measuring the paraphrasing capacity and can't distinguish style.  

\begin{table*}[h]
\centering
\caption{Results \texttt{Meta-Llama-3-8B-Instruct} across all methods and users.} 
\label{tab:results_llama3}
\scalebox{0.55}{
\renewcommand{\arraystretch}{1.20}

\begin{adjustbox}{center}

\begin{tabular}{l>{\columncolor{gray!10}}c>{\columncolor{gray!10}}c>{\columncolor{gray!10}}cccc>{\columncolor{gray!10}}c>{\columncolor{gray!10}}c>{\columncolor{gray!10}}cccc>{\columncolor{gray!10}}c>{\columncolor{gray!10}}c>{\columncolor{gray!10}}c}%
\hline%
 &\multicolumn{3}{c}{\texttt{David}}&\multicolumn{3}{c}{\texttt{Jeff}}&\multicolumn{3}{c}{\texttt{Kay}}&\multicolumn{3}{c}{\texttt{Sara}}&\multicolumn{3}{c}{\texttt{Tana}}\\%
\hline%
Method&BLEU&Rouge&MAUVE&BLEU&Rouge&MAUVE&BLEU&Rouge&MAUVE&BLEU&Rouge&MAUVE&BLEU&Rouge&MAUVE\\%
\texttt{Pretrained}&$0.083$&$0.181$&$0.009$&$0.108$&$0.182$&$0.004$&$0.113$&$0.186$&$0.005$&$0.144$&$0.23$&$0.004$&$0.132$&$0.21$&$0.006$\\%
\texttt{Pretrained{-}RAG}&$0.107$&$0.212$&$0.017$&$0.115$&$0.188$&$0.005$&$0.121$&$0.197$&$0.004$&$0.151$&$0.233$&$0.004$&$0.149$&$0.227$&$0.005$\\%
\hline%
\texttt{FFT}&$0.278$&$0.46$&$0.996$&$0.166$&$0.282$&$0.758$&$0.197$&$\textbf{0.295}$&$0.863$&$0.261$&$0.356$&$0.859$&$0.256$&$0.358$&$0.859$\\%
\texttt{FFT{-}RAG}&$0.3$&$0.449$&$0.984$&$0.166$&$0.266$&$0.779$&$0.19$&$0.283$&$0.933$&$0.242$&$0.337$&$0.852$&$0.238$&$0.327$&$0.898$\\%
\texttt{FFT{-}RAFT}&$0.299$&$0.476$&$0.997$&$0.164$&$0.279$&$0.715$&$0.192$&$0.285$&$0.891$&$0.253$&$0.357$&$0.914$&$0.266$&$0.363$&$0.903$\\%
\texttt{FFT{-}RAFT{-}RAG}&$0.31$&$0.494$&$0.985$&$0.187$&$0.297$&$0.824$&$0.184$&$0.281$&$0.941$&$0.263$&$0.36$&$0.826$&$\textbf{0.278}$&$\textbf{0.372}$&$0.876$\\%
\hline%
\texttt{RoSA}&$0.312$&$0.488$&$0.999$&$0.164$&$0.285$&$0.806$&$\textbf{0.202}$&$0.293$&$0.898$&$0.26$&$0.355$&$0.872$&$0.26$&$0.352$&$0.948$\\%
\texttt{RoSA{-}RAG}&$0.236$&$0.383$&$0.982$&$0.166$&$0.266$&$0.786$&$0.184$&$0.269$&$0.963$&$0.243$&$0.338$&$0.945$&$0.253$&$0.34$&$\textbf{0.97}$\\%
\texttt{RoSA{-}RAFT}&$0.321$&$0.491$&$0.991$&$0.166$&$0.291$&$0.823$&$0.201$&$0.29$&$0.895$&$\textbf{0.268}$&$0.36$&$0.763$&$0.252$&$0.349$&$0.814$\\%
\texttt{RoSA{-}RAFT{-}RAG}&$\textbf{0.346}$&$\textbf{0.509}$&$\textbf{1.0}$&$\textbf{0.192}$&$\textbf{0.305}$&$\textbf{0.867}$&$0.196$&$0.291$&$0.973$&$0.265$&$\textbf{0.362}$&$0.889$&$0.277$&$0.367$&$0.905$\\%
\hline%
\texttt{LoRA}&$0.26$&$0.401$&$0.68$&$0.167$&$0.27$&$0.224$&$0.184$&$0.269$&$0.979$&$0.247$&$0.351$&$0.921$&$0.239$&$0.329$&$0.897$\\%
\texttt{LoRA{-}RAG}&$0.257$&$0.406$&$0.922$&$0.159$&$0.253$&$0.481$&$0.181$&$0.267$&$0.968$&$0.23$&$0.327$&$\textbf{0.969}$&$0.233$&$0.312$&$0.942$\\%
\texttt{LoRA{-}RAFT}&$0.243$&$0.391$&$0.412$&$0.165$&$0.267$&$0.149$&$0.187$&$0.276$&$\textbf{0.987}$&$0.245$&$0.341$&$0.906$&$0.243$&$0.329$&$0.893$\\%
\texttt{LoRA{-}RAFT{-}RAG}&$0.246$&$0.404$&$0.899$&$0.167$&$0.262$&$0.547$&$0.185$&$0.27$&$0.942$&$0.252$&$0.349$&$0.909$&$0.262$&$0.353$&$0.854$\\%
\hline%
\end{tabular}%

\end{adjustbox}

}
\end{table*}

\begin{table*}[h]
\centering
\caption{Results \texttt{Mistral-7B-Instruct-v0.2} across all methods and users.}
\label{tab:results_mistral_v2}
\scalebox{0.55}{
\renewcommand{\arraystretch}{1.20}

\begin{adjustbox}{center}

\begin{tabular}{l>{\columncolor{gray!10}}c>{\columncolor{gray!10}}c>{\columncolor{gray!10}}cccc>{\columncolor{gray!10}}c>{\columncolor{gray!10}}c>{\columncolor{gray!10}}cccc>{\columncolor{gray!10}}c>{\columncolor{gray!10}}c>{\columncolor{gray!10}}c}%
\hline%
 &\multicolumn{3}{c}{\texttt{David}}&\multicolumn{3}{c}{\texttt{Jeff}}&\multicolumn{3}{c}{\texttt{Kay}}&\multicolumn{3}{c}{\texttt{Sara}}&\multicolumn{3}{c}{\texttt{Tana}}\\%
\hline%
Method&BLEU&Rouge&MAUVE&BLEU&Rouge&MAUVE&BLEU&Rouge&MAUVE&BLEU&Rouge&MAUVE&BLEU&Rouge&MAUVE\\%
\hline
\texttt{Pretrained}&$0.072$&$0.158$&$0.009$&$0.093$&$0.164$&$0.005$&$0.096$&$0.168$&$0.006$&$0.117$&$0.196$&$0.004$&$0.113$&$0.187$&$0.007$\\%
\texttt{Pretrained{-}RAG}&$0.083$&$0.181$&$0.058$&$0.102$&$0.17$&$0.006$&$0.103$&$0.176$&$0.017$&$0.122$&$0.2$&$0.008$&$0.126$&$0.201$&$0.005$\\%
\hline%
\texttt{FFT}&$\textbf{0.335}$&$\textbf{0.475}$&$0.997$&$0.162$&$0.262$&$0.263$&$0.214$&$0.295$&$0.795$&$0.264$&$0.365$&$0.839$&$0.263$&$0.36$&$0.719$\\%
\texttt{FFT{-}RAG}&$0.276$&$0.425$&$0.934$&$0.162$&$0.251$&$0.376$&$0.2$&$0.286$&$0.896$&$0.236$&$0.335$&$0.876$&$0.255$&$0.346$&$\textbf{0.872}$\\%
\texttt{FFT{-}RAFT}&$0.306$&$0.461$&$0.963$&$0.163$&$0.262$&$0.346$&$0.211$&$\textbf{0.313}$&$0.696$&$\textbf{0.28}$&$\textbf{0.393}$&$0.792$&$0.254$&$0.35$&$0.759$\\%
\texttt{FFT{-}RAFT{-}RAG}&$0.309$&$0.461$&$0.959$&$\textbf{0.174}$&$\textbf{0.277}$&$0.29$&$\textbf{0.216}$&$0.308$&$0.882$&$0.275$&$0.39$&$0.648$&$0.268$&$0.358$&$0.723$\\%
\hline%
\texttt{RoSA}&$0.306$&$0.459$&$0.988$&$0.169$&$0.258$&$0.482$&$0.209$&$0.302$&$0.797$&$0.253$&$0.358$&$0.955$&$0.24$&$0.33$&$0.74$\\%
\texttt{RoSA{-}RAG}&$0.263$&$0.416$&$\textbf{1.0}$&$0.154$&$0.228$&$0.517$&$0.182$&$0.259$&$0.894$&$0.2$&$0.287$&$\textbf{0.971}$&$0.238$&$0.331$&$0.826$\\%
\texttt{RoSA{-}RAFT}&$0.289$&$0.437$&$0.999$&$0.167$&$0.264$&$0.557$&$0.189$&$0.296$&$0.809$&$0.265$&$0.387$&$0.838$&$0.245$&$0.337$&$0.856$\\%
\texttt{RoSA{-}RAFT{-}RAG}&$0.289$&$0.43$&$0.998$&$0.171$&$0.262$&$0.408$&$0.204$&$0.293$&$\textbf{0.961}$&$0.255$&$0.37$&$0.827$&$\textbf{0.273}$&$\textbf{0.365}$&$0.757$\\%
\hline%
\texttt{LoRA}&$0.294$&$0.444$&$0.976$&$0.157$&$0.247$&$0.56$&$0.145$&$0.229$&$0.44$&$0.214$&$0.318$&$0.413$&$0.214$&$0.318$&$0.288$\\%
\texttt{LoRA{-}RAG}&$0.249$&$0.397$&$0.974$&$0.166$&$0.25$&$0.589$&$0.169$&$0.257$&$0.891$&$0.234$&$0.336$&$0.702$&$0.241$&$0.336$&$0.535$\\%
\texttt{LoRA{-}RAFT}&$0.298$&$0.451$&$0.941$&$0.158$&$0.248$&$\textbf{0.681}$&$0.136$&$0.217$&$0.426$&$0.204$&$0.3$&$0.465$&$0.206$&$0.303$&$0.3$\\%
\texttt{LoRA{-}RAFT{-}RAG}&$0.286$&$0.437$&$0.991$&$0.168$&$0.262$&$0.569$&$0.151$&$0.235$&$0.417$&$0.23$&$0.334$&$0.471$&$0.232$&$0.329$&$0.329$\\%
\hline%
\end{tabular}

\end{adjustbox}

}
\end{table*}

\begin{table*}[h!]
\centering
\caption{Results for \texttt{Phi-3-mini-4k-instruct} across all methods and users.}
\label{tab:results_phi3}
\scalebox{0.55}{
\renewcommand{\arraystretch}{1.20}

\begin{adjustbox}{center}

\normalsize%
\begin{tabular}{l>{\columncolor{gray!10}}c>{\columncolor{gray!10}}c>{\columncolor{gray!10}}cccc>{\columncolor{gray!10}}c>{\columncolor{gray!10}}c>{\columncolor{gray!10}}cccc>{\columncolor{gray!10}}c>{\columncolor{gray!10}}c>{\columncolor{gray!10}}c}%
\hline%
 &\multicolumn{3}{c}{\texttt{David}}&\multicolumn{3}{c}{\texttt{Jeff}}&\multicolumn{3}{c}{\texttt{Kay}}&\multicolumn{3}{c}{\texttt{Sara}}&\multicolumn{3}{c}{\texttt{Tana}}\\%
\hline%
Method&BLEU&Rouge&MAUVE&BLEU&Rouge&MAUVE&BLEU&Rouge&MAUVE&BLEU&Rouge&MAUVE&BLEU&Rouge&MAUVE\\%
\hline
\texttt{Pretrained}&$0.055$&$0.129$&$0.009$&$0.077$&$0.147$&$0.006$&$0.078$&$0.149$&$0.006$&$0.101$&$0.182$&$0.007$&$0.093$&$0.168$&$0.005$\\%
\texttt{Pretrained{-}RAG}&$0.054$&$0.126$&$0.009$&$0.082$&$0.151$&$0.007$&$0.082$&$0.151$&$0.012$&$0.107$&$0.185$&$0.009$&$0.101$&$0.176$&$0.009$\\%
\hline%
\texttt{FFT}&$0.33$&$0.488$&$1.0$&$0.169$&$0.281$&$0.86$&$0.199$&$0.299$&$0.866$&$0.27$&$0.379$&$0.884$&$0.265$&$0.364$&$0.869$\\%
\texttt{FFT{-}RAG}&$0.31$&$0.474$&$0.994$&$0.179$&$0.28$&$0.726$&$0.196$&$0.283$&$0.858$&$0.231$&$0.322$&$0.959$&$0.254$&$0.353$&$0.911$\\%
\texttt{FFT{-}RAFT}&$\textbf{0.33}$&$\textbf{0.508}$&$0.992$&$0.166$&$0.276$&$\textbf{0.915}$&$0.194$&$0.289$&$0.923$&$0.261$&$0.366$&$0.881$&$0.262$&$0.371$&$0.877$\\%
\texttt{FFT{-}RAFT{-}RAG}&$0.327$&$0.495$&$0.998$&$\textbf{0.181}$&$0.289$&$0.824$&$0.201$&$0.294$&$0.862$&$0.263$&$0.365$&$0.865$&$\textbf{0.276}$&$\textbf{0.38}$&$0.887$\\%
\hline%
\texttt{RoSA}&$0.308$&$0.473$&$0.996$&$0.166$&$0.256$&$0.85$&$0.193$&$0.281$&$\textbf{0.956}$&$0.241$&$0.35$&$0.866$&$0.245$&$0.337$&$0.866$\\%
\texttt{RoSA{-}RAG}&$0.302$&$0.455$&$1.0$&$0.171$&$0.267$&$0.726$&$0.171$&$0.256$&$0.814$&$0.195$&$0.289$&$\textbf{0.982}$&$0.251$&$0.339$&$0.948$\\%
\texttt{RoSA{-}RAFT}&$0.315$&$0.473$&$\textbf{1.0}$&$0.157$&$0.257$&$0.875$&$0.18$&$0.278$&$0.916$&$0.234$&$0.343$&$0.939$&$0.245$&$0.331$&$\textbf{0.958}$\\%
\texttt{RoSA{-}RAFT{-}RAG}&$0.319$&$0.472$&$0.996$&$0.173$&$0.274$&$0.869$&$0.188$&$0.278$&$0.896$&$0.238$&$0.345$&$0.812$&$0.265$&$0.354$&$0.945$\\%
\hline%
\texttt{LoRA}&$0.277$&$0.439$&$1.0$&$0.164$&$0.289$&$0.789$&$0.2$&$0.302$&$0.72$&$0.261$&$0.384$&$0.754$&$0.253$&$0.359$&$0.82$\\%
\texttt{LoRA{-}RAG}&$0.295$&$0.461$&$0.999$&$0.173$&$0.283$&$0.77$&$0.156$&$0.246$&$0.577$&$0.235$&$0.353$&$0.909$&$0.25$&$0.347$&$0.834$\\%
\texttt{LoRA{-}RAFT}&$0.288$&$0.443$&$1.0$&$0.16$&$0.286$&$0.846$&$\textbf{0.21}$&$\textbf{0.319}$&$0.852$&$\textbf{0.276}$&$\textbf{0.398}$&$0.767$&$0.261$&$0.366$&$0.802$\\%
\texttt{LoRA{-}RAFT{-}RAG}&$0.313$&$0.472$&$0.999$&$0.176$&$\textbf{0.3}$&$0.65$&$0.209$&$0.311$&$0.919$&$0.269$&$0.389$&$0.923$&$0.274$&$0.376$&$0.757$\\%
\hline%
\end{tabular}%

\end{adjustbox}

}
\end{table*}

\begin{figure*}[h]
  \centering
  \renewcommand{\tabcolsep}{0.4em}
  \renewcommand{\arraystretch}{1}
  \resizebox{\textwidth}{!}{
  \begin{tabular}{cc}
    ~&\hspace*{0.15in}\scriptsize MAUVE\hspace*{1.90in}\scriptsize BLEU \hspace*{1.81in} \scriptsize ROUGE \\
    \raisebox{0.95in}{\rotatebox{90}{\scriptsize\textsf{User train}}}&\hspace*{-1mm}
    \includegraphics[height=.33\linewidth]{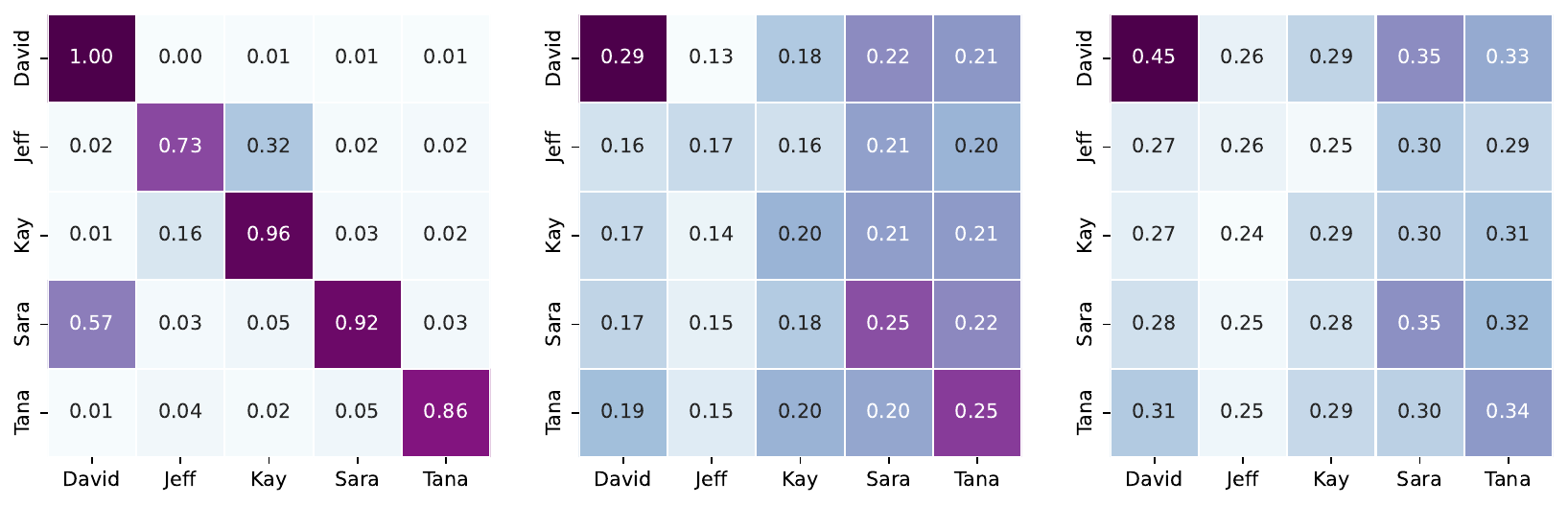}\\[-2.5pt]
    ~&\hspace*{0.15in}\scriptsize\textsf{User test}\\
  \end{tabular}\vspace{-5pt}
  }
  \caption{\label{figConfusionMatrices}
  Style comparison between  \texttt{Phi-3-mini-4k-\-instruct} models trained for different users. 
  }
\end{figure*}

\begin{table*}[tp]
\centering
\caption{Comparison between models (results average over all users).}
\label{tab:results_all_models}
\scalebox{0.70}{
\renewcommand{\arraystretch}{1.2}

\begin{adjustbox}{center}

\begin{tabular}{l>{\columncolor{gray!10}}c>{\columncolor{gray!10}}c>{\columncolor{gray!10}}cccc>{\columncolor{gray!10}}c>{\columncolor{gray!10}}c>{\columncolor{gray!10}}c}%
\hline%
 &\multicolumn{3}{c}{\texttt{Mistral}}&\multicolumn{3}{c}{\texttt{Llama-3}}&\multicolumn{3}{c}{\texttt{Phi-3}}\\%
\hline%
Method&BLEU&Rouge&MAUVE&BLEU&Rouge&MAUVE&BLEU&Rouge&MAUVE\\%
\hline
\texttt{Pretrained}&$0.098$&$0.174$&$0.006$&$0.116$&$0.198$&$0.006$&$0.081$&$0.155$&$0.007$\\%
\texttt{Pretrained{-}RAG}&$0.107$&$0.186$&$0.019$&$0.129$&$0.212$&$0.007$&$0.085$&$0.158$&$0.009$\\%
\hline%
\texttt{FFT}&$0.247$&$0.351$&$0.722$&$0.231$&$0.35$&$0.867$&$0.247$&$0.362$&$0.896$\\%
\texttt{FFT{-}RAG}&$0.226$&$0.328$&$0.791$&$0.227$&$0.332$&$0.889$&$0.234$&$0.342$&$0.89$\\%
\texttt{FFT{-}RAFT}&$0.243$&$0.356$&$0.711$&$0.235$&$0.352$&$0.884$&$0.243$&$0.362$&$0.918$\\%
\texttt{FFT{-}RAFT{-}RAG}&$\textbf{0.248}$&$\textbf{0.359}$&$0.7$&$0.244$&$0.361$&$0.89$&$\textbf{0.25}$&$0.365$&$0.887$\\%
\hline%
\texttt{RoSA}&$0.235$&$0.341$&$0.792$&$0.24$&$0.355$&$0.904$&$0.231$&$0.339$&$0.907$\\%
\texttt{RoSA{-}RAG}&$0.208$&$0.304$&$\textbf{0.841}$&$0.216$&$0.319$&$\textbf{0.929}$&$0.218$&$0.321$&$0.894$\\%
\texttt{RoSA{-}RAFT}&$0.231$&$0.344$&$0.812$&$0.241$&$0.356$&$0.857$&$0.226$&$0.336$&$\textbf{0.938}$\\%
\texttt{RoSA{-}RAFT{-}RAG}&$0.238$&$0.344$&$0.79$&$\textbf{0.255}$&$\textbf{0.367}$&$0.927$&$0.237$&$0.345$&$0.904$\\%
\hline%
\texttt{LoRA}&$0.205$&$0.311$&$0.535$&$0.219$&$0.324$&$0.74$&$0.231$&$0.355$&$0.817$\\%
\texttt{LoRA{-}RAG}&$0.212$&$0.315$&$0.738$&$0.212$&$0.313$&$0.856$&$0.222$&$0.338$&$0.818$\\%
\texttt{LoRA{-}RAFT}&$0.2$&$0.304$&$0.563$&$0.216$&$0.321$&$0.669$&$0.239$&$0.363$&$0.853$\\%
\texttt{LoRA{-}RAFT{-}RAG}&$0.213$&$0.319$&$0.556$&$0.222$&$0.328$&$0.83$&$0.248$&$\textbf{0.37}$&$0.849$\\%
\hline%
\end{tabular}

\end{adjustbox}

}
\end{table*}

\begin{table*}[h]
\centering
\caption{Anon users results (\texttt{Meta-Llama-3-8B-Instruct})}
\label{tab:results_llama3_anon}
\scalebox{0.70}{
\renewcommand{\arraystretch}{1.2}

\begin{adjustbox}{center}

\begin{tabular}{l>{\columncolor{gray!10}}c>{\columncolor{gray!10}}c>{\columncolor{gray!10}}cccc}%
\hline%
 &\multicolumn{3}{c}{\texttt{Anon1}}&\multicolumn{3}{c}{\texttt{Anon2}}\\%
\hline%
Method&BLEU&Rouge&Mauve&BLEU&Rouge&Mauve\\%
\hline
\texttt{Pretrained}&$0.11$&$0.207$&$0.006$&$0.094$&$0.18$&$0.005$\\%
\texttt{Pretrained{-}RAG}&$0.123$&$0.226$&$0.006$&$0.108$&$0.197$&$0.007$\\%
\hline%
\texttt{FFT}&$0.302$&$0.462$&$0.889$&$0.219$&$0.381$&$0.909$\\%
\texttt{FFT{-}RAG}&$0.232$&$0.398$&$0.918$&$0.167$&$0.301$&$\textbf{0.976}$\\%
\texttt{FFT{-}RAFT}&$0.293$&$\textbf{0.464}$&$0.959$&$0.214$&$\textbf{0.383}$&$0.898$\\%
\texttt{FFT{-}RAFT{-}RAG}&$0.287$&$0.461$&$0.95$&$0.218$&$0.369$&$0.895$\\%
\hline%
\texttt{RoSA}&$\textbf{0.306}$&$0.459$&$0.958$&$\textbf{0.22}$&$0.358$&$0.927$\\%
\texttt{RoSA{-}RAG}&$0.278$&$0.435$&$0.975$&$0.198$&$0.327$&$0.934$\\%
\texttt{RoSA{-}RAFT}&$0.285$&$0.436$&$0.978$&$0.21$&$0.348$&$0.951$\\%
\texttt{RoSA{-}RAFT{-}RAG}&$0.29$&$0.438$&$0.961$&$0.217$&$0.353$&$0.939$\\%
\hline%
\texttt{LoRA}&$0.3$&$0.457$&$0.959$&$0.208$&$0.348$&$0.96$\\%
\texttt{LoRA{-}RAG}&$0.245$&$0.389$&$0.95$&$0.2$&$0.323$&$0.954$\\%
\texttt{LoRA{-}RAFT}&$0.289$&$0.45$&$0.965$&$0.206$&$0.349$&$0.956$\\%
\texttt{LoRA{-}RAFT{-}RAG}&$0.288$&$0.443$&$\textbf{0.985}$&$0.214$&$0.353$&$0.957$\\%
\hline%
\end{tabular}

\end{adjustbox}

}
\end{table*}

\begin{table*}[h]
\centering
\caption{Anon users results (\texttt{Mistral-7B-Instruct-v0.2})}
\label{tab:results_mistral2_anon}
\scalebox{0.70}{
\renewcommand{\arraystretch}{1.2}

\begin{adjustbox}{center}

\begin{tabular}{l>{\columncolor{gray!10}}c>{\columncolor{gray!10}}c>{\columncolor{gray!10}}cccc}%
\hline%
 &\multicolumn{3}{c}{\texttt{Anon1}}&\multicolumn{3}{c}{\texttt{Anon2}}\\%
\hline%
Method&BLEU&Rouge&Mauve&BLEU&Rouge&Mauve\\%
\hline
\texttt{Pretrained}&$0.085$&$0.173$&$0.011$&$0.081$&$0.16$&$0.005$\\%
\texttt{Pretrained{-}RAG}&$0.095$&$0.188$&$0.011$&$0.089$&$0.169$&$0.008$\\%
\hline%
\texttt{FFT}&$0.291$&$\textbf{0.459}$&$0.878$&$0.214$&$0.362$&$0.706$\\%
\texttt{FFT{-}RAG}&$0.248$&$0.411$&$0.824$&$0.197$&$0.334$&$0.812$\\%
\texttt{FFT{-}RAFT}&$0.295$&$0.457$&$0.877$&$0.217$&$\textbf{0.367}$&$0.559$\\%
\texttt{FFT{-}RAFT{-}RAG}&$0.287$&$0.45$&$0.825$&$\textbf{0.222}$&$0.362$&$0.679$\\%
\hline%
\texttt{RoSA}&$0.294$&$0.459$&$0.912$&$0.213$&$0.35$&$0.888$\\%
\texttt{RoSA{-}RAG}&$0.266$&$0.422$&$0.875$&$0.175$&$0.292$&$0.914$\\%
\texttt{RoSA{-}RAFT}&$\textbf{0.298}$&$0.458$&$0.851$&$0.215$&$0.347$&$0.868$\\%
\texttt{RoSA{-}RAFT{-}RAG}&$0.286$&$0.448$&$0.907$&$0.206$&$0.336$&$0.832$\\%
\hline%
\texttt{LoRA}&$0.275$&$0.415$&$0.935$&$0.203$&$0.335$&$\textbf{0.948}$\\%
\texttt{LoRA{-}RAG}&$0.259$&$0.401$&$0.943$&$0.189$&$0.314$&$0.93$\\%
\texttt{LoRA{-}RAFT}&$0.28$&$0.43$&$0.96$&$0.203$&$0.335$&$0.916$\\%
\texttt{LoRA{-}RAFT{-}RAG}&$0.278$&$0.429$&$\textbf{0.975}$&$0.206$&$0.335$&$0.901$\\%
\hline%
\end{tabular}

\end{adjustbox}

}
\end{table*}

\begin{table*}[h]
\centering
\caption{Anon users results (\texttt{Phi-3-mini-4k-instruct})}
\label{tab:results_phi3_anon}
\scalebox{0.70}{
\renewcommand{\arraystretch}{1.2}

\begin{adjustbox}{center}

\begin{tabular}{l>{\columncolor{gray!10}}c>{\columncolor{gray!10}}c>{\columncolor{gray!10}}cccc}%
\hline%
 &\multicolumn{3}{c}{\texttt{Anon1}}&\multicolumn{3}{c}{\texttt{Anon2}}\\%
\hline%
Method&BLEU&Rouge&Mauve&BLEU&Rouge&Mauve\\%
\hline
\texttt{Pretrained}&$0.067$&$0.144$&$0.009$&$0.069$&$0.139$&$0.006$\\%
\texttt{Pretrained{-}RAG}&$0.067$&$0.145$&$0.01$&$0.068$&$0.139$&$0.007$\\%
\hline%
\texttt{FFT}&$\textbf{0.306}$&$0.468$&$0.914$&$0.23$&$0.383$&$0.942$\\%
\texttt{FFT{-}RAG}&$0.302$&$0.462$&$0.888$&$0.219$&$0.359$&$0.939$\\%
\texttt{FFT{-}RAFT}&$0.303$&$\textbf{0.477}$&$0.968$&$\textbf{0.233}$&$\textbf{0.394}$&$0.943$\\%
\texttt{FFT{-}RAFT{-}RAG}&$0.3$&$0.473$&$0.923$&$0.229$&$0.374$&$0.943$\\%
\hline%
\texttt{RoSA}&$0.293$&$0.446$&$0.926$&$0.206$&$0.356$&$0.924$\\%
\texttt{RoSA{-}RAG}&$0.3$&$0.457$&$0.927$&$0.21$&$0.348$&$0.927$\\%
\texttt{RoSA{-}RAFT}&$0.288$&$0.449$&$\textbf{0.98}$&$0.213$&$0.358$&$0.941$\\%
\texttt{RoSA{-}RAFT{-}RAG}&$0.284$&$0.445$&$0.907$&$0.215$&$0.35$&$0.926$\\%
\hline%
\texttt{LoRA}&$0.296$&$0.468$&$0.945$&$0.22$&$0.38$&$\textbf{0.95}$\\%
\texttt{LoRA{-}RAG}&$0.292$&$0.454$&$0.886$&$0.223$&$0.373$&$0.935$\\%
\texttt{LoRA{-}RAFT}&$0.302$&$0.467$&$0.951$&$0.217$&$0.386$&$0.885$\\%
\texttt{LoRA{-}RAFT{-}RAG}&$0.301$&$0.466$&$0.945$&$0.226$&$0.379$&$0.924$\\%
\hline%
\end{tabular}

\end{adjustbox}

}
\end{table*}

\section{Hyperparameter Tuning}
\label{sec:appendixHyperparameters}

\subsection{Inference}

We perform generation using beam search with a temperature $T=0.7$, number of top probability tokens to keep $top\_k = 50$ and nucleus sampling parameter $top\_p=0.7$.

\subsection{Fine-Tuning}

We perform a thorough hyperparameter tuning for every backbone, and every user over learning rate and number of epochs. For FFT we experiment with \{$1$, $3$, $5$\} epochs and learning rates between $[10^{-3}, 10^{-7}]$. For PEFT methods (RoSA, LoRA), we experiment with \{$1$, $3$, $5$, $7$, $9$\} epochs and learning rates between $[10^{-3}, 10^{-7}]$. Next, we present the best configuration found for each model, method and user. 

\paragraph{Meta-Llama-3-8B-Instruct} For users David and Jeff: FFT for $3$ epochs with a learning rate of $10^{-5}$; PEFT for $7$ epochs with a learning rate of $10^{-5}$. For users Kay, Sara and Tana: FFT for $3$ epochs with a learning rate of $10^{-4}$ and PEFT for $7$ epochs with a learning rate of $10^{-4}$. For users Anon1, Anon2: FFT for $3$ epochs with a learning rate of $10^{-5}$ and PEFT for $7$ epochs with a learning rate of $10^{-4}$.
\paragraph{Mistral-7B-Instruct-v0.2} For users David and Jeff, Anon2: FFT for $3$ epochs with a learning rate of $10^{-5}$; PEFT for $7$ epochs with a learning rate of $10^{-4}$. For users Kay, Sara and Tana: FFT for $3$ epochs with a learning rate of $10^{-5}$ and PEFT for $7$ epochs with a learning rate of $10^{-5}$. For Anon1: FFT for $3$ epochs with a learning rate of $10^{-5}$; RoSA for $7$ epochs with a learning rate of $10^{-5}$, LoRA for $7$ epochs with a learning rate of $10^{-4}$.
\paragraph{Phi-3-mini-4k-instruct} Same hyperparameters for every users: FFT for $3$ epochs with a learning rate of $10^{-4}$; PEFT for $7$ epochs wth a learning rate of $10^{-4}$.

\subsection{RAG and RAFT}
We set fixed values of the number of retrieval emails $n_{RAG} = 2$ during RAFT, and $n_{RAG} = 3$ at inference with RAG. We pick relatively low values, as every email retrieved increases the input length, thus the memory consumption. We experimented with larger $n_{RAG}$ for the Pretrained baseline, without significantly different results. For RAFT, we use $p_{RAG} = 0.55$ chance to include relevant emails in the prompt. We use the same relevancy threshold $T_{RAG} = 0.2$, tuned on the private users Anon1 and Anon2 to encourage recall rather than precision. This can retrieve irrelevant emails, but prevents missing important items from the user's history. Furthermore, RAFT learns how to better filter irrelevant information in case it is added to the prompt.

\end{document}